\documentclass[lettersize,journal]{IEEEtran}
\usepackage{amsmath,amsfonts}
\usepackage{algorithmic}
\usepackage{algorithm}
\usepackage{array}
\usepackage[caption=false,font=normalsize,labelfont=sf,textfont=sf]{subfig}
\usepackage{textcomp}
\usepackage{stfloats}
\usepackage{url}
\usepackage{verbatim}
\usepackage{graphicx}
\usepackage{cite}
\usepackage{multirow}
\usepackage{mathrsfs}
\usepackage{color}
\usepackage{hyperref}
\hypersetup{
	colorlinks=true,
	linkcolor=blue,
	filecolor=blue,      
	urlcolor=blue,
	citecolor=blue,
}
\begin{document}

\title{Advancing Compositional LLM Reasoning with Structured Task Relations in Interactive Multimodal Communications}

\author{Xinye Cao,\IEEEmembership{~Graduate Student Member,~IEEE,} Hongcan Guo, Guoshun Nan,\IEEEmembership{~Member,~IEEE,} Jiaoyang Cui, Haoting Qian, Yihan Lin, Yilin Peng, Diyang Zhang, Yanzhao Hou,\IEEEmembership{~Member,~IEEE,} Huici Wu,\IEEEmembership{~Member,~IEEE,} Xiaofeng Tao,\IEEEmembership{~Senior Member,~IEEE,} Tony Q.S. Quek,\IEEEmembership{~Fellow,~IEEE}

\thanks{This work was supported in part by the National Key Research and Development Program of China under Grant 2022YFB2902200; in part by the National Natural Science Foundation of China under Grant 62471064; in part by Beijing University of Posts and Telecommunications (BUPT) Excellent Ph. D. Students Foundation under Grant CX20252013; in part by the National Research Foundation, Singapore and Infocomm Media Development Authority under its Future Communications Research \& Development Programme; and in part by the SNS JU project 6G-GOALS under the EU’s Horizon program Grant Agreement No. 101139232. (Corresponding author: Guoshun Nan.)}
\thanks{X. Cao, H. Guo, G. Nan, J. Cui, H. Qian, Y. Lin, Y. Peng, D. Zhang, Y. Hou, H. Wu, X. Tao are with National Engineering Research Center for Mobile Network Technologies, Beijing University of Posts and Telecommunications, Beijing 100876, China and BUPT Shenzhen Institute, Shenzhen, China. (e-mail: caoxinye@bupt.edu.cn; ai.guohc@bupt.edu.cn; nanguo2021@bupt.edu.cn; cuijiaoyang24@bupt.edu.cn; Qian\_Haoting@bupt.edu.cn; LinJHS@bupt.edu.cn; Peng\_yl@bupt.edu.cn; midorizhang@bupt.edu.cn; houyanzhao@bupt.edu.cn; dailywu@bupt.edu.cn; taoxf.bupt@gmail.com).}
\thanks{T. Q. S. Quek is with the Singapore University of Technology and Design, Singapore 487372, and also with the Yonsei Frontier Lab, Yonsei University, South Korea (e-mail: tonyquek@sutd.edu.sg).}
\thanks{Xinye Cao and Hongcan Guo are equally contributed.}
\thanks{Jiaoyang Cui and Haoting Qian are equally contributed.}
\thanks{This work has been submitted to the IEEE for possible publication. Copyright may be transferred without notice, after which this version may no longer be accessible.}
}



\maketitle

\begin{abstract}
Interactive multimodal applications (IMAs), such as route planning in the Internet of Vehicles, enrich users' personalized experiences by integrating various forms of data over wireless networks. Recent advances in large language models (LLMs) utilize mixture-of-experts (MoE) mechanisms to empower multiple IMAs, with each LLM trained individually for a specific task that presents different business workflows. In contrast to existing approaches that rely on multiple LLMs for IMAs, this paper presents a novel paradigm that accomplishes various IMAs using a single compositional LLM over wireless networks. The two primary challenges include 1) guiding a single LLM to adapt to diverse IMA objectives and 2) ensuring the flexibility and efficiency of the LLM in resource-constrained mobile environments. To tackle the first challenge, we propose ContextLoRA, a novel method that guides an LLM to learn the rich structured context among IMAs by constructing a task dependency graph. We partition the learnable parameter matrix of neural layers for each IMA to facilitate LLM composition. Then, we develop a step-by-step fine-tuning procedure guided by task relations, including training, freezing, and masking phases. This allows the LLM to learn to reason among tasks for better adaptation, capturing the latent dependencies between tasks. For the second challenge, we introduce ContextGear, a scheduling strategy to optimize the training procedure of ContextLoRA, aiming to minimize computational and communication costs through a strategic grouping mechanism. Experiments on three benchmarks show the superiority of the proposed ContextLoRA and ContextGear. Furthermore, we prototype our proposed paradigm on a real-world wireless testbed, demonstrating its practical applicability for various IMAs. We will release our code to the community.
\end{abstract}

\begin{IEEEkeywords}
Interactive multimodal communication, large language models, multi-task learning.
\end{IEEEkeywords}

\section{Introduction}
\label{1Intro}
\subsection{Background}

\begin{figure}[!t]
\centering
\includegraphics[width=0.47\textwidth,trim=15 80 405 0,clip]{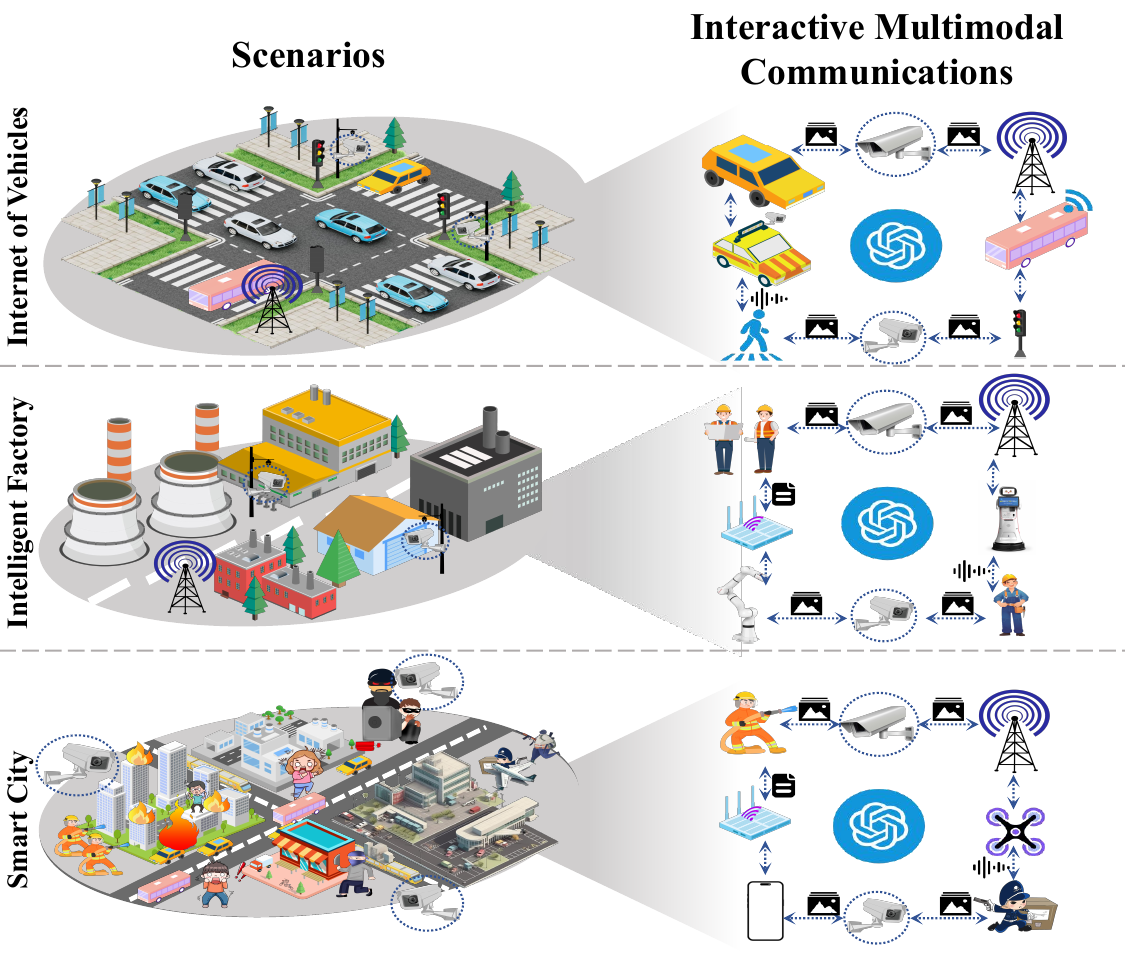}
\caption{Interactive multimodal applications (IMAs) over wireless networks, such as the Internet of Vehicles, intelligent factories, and smart cities. Users engage with IMAs using various multimodal inputs, such as voice, text, and images, while these IMAs generate intelligent decisions tailored to requests.}
\label{fig_scenario}
\vspace{-3mm}
\end{figure}

\IEEEPARstart{I}{nteractive} multimodal applications (IMAs)~\cite{jabeen2023review}, such as route planning in the Internet of Vehicles~\cite{wen2024road,lv20206g,liu2024llm,10013691} and anomaly detection in smart cities, expand the landscape of intelligent communications by leveraging cross-modal data over wireless networks. IMAs also have great potential in the sixth generation of communications (6G)~\cite{ITU6G,9078581,sui2023mining}, facilitating 6G usage scenarios such as immersive communications, and artificial intelligence and communication~\cite{she2021tutorial,9390169,8808168,cui2025overview} remarked by the International Telecommunication Union (ITU). Figure \ref{fig_scenario} showcases three scenarios of IMAs, including the Internet of Vehicles (IoV)~\cite{IEEE2020IoV,10654917,10013691}, intelligent factories (IF)~\cite{wang2024llm,SCIE1_2020SmartFac}, and smart cities (SC)~\cite{see2018artificial}. For example, in intelligent factories, a user can remotely interact with a robot to take actions, such as moving boxes and operating assembly lines, and robots can also collaborate through interactions over wireless networks. These models use various multimodal inputs, such as voice, text, and visual cues, to make intelligent decisions for users and devices during interactions. Early studies~\cite{mohammadi2017semisupervised, Du2024MRL,Tao2021MRL,Jiao2024MRL,Zhang2023MRL} mainly rely on reinforcement learning to help automate such a procedure. The key issue of these methods lies in the generalization of a specific model across various tasks.

\subsection{Motivation}
Recently proliferated LLMs~\cite{10386743, wunext, wang2021survey} hold the promise of promoting diverse IMAs. A pre-trained foundation LLM can be further tuned to various downstream tasks~\cite{huang2023reparameterized}. Along this line, low-rank adaptation (LoRA)~\cite{hulora} is a well-known technique to fine-tune LLMs for specific tasks without retraining the entire model. Additionally, the popular mixture-of-experts (MoE) mechanisms~\cite{NEURIPS2022MoE} based on LoRA can facilitate multiple IMAs, with each LLM being trained independently for a specific task that corresponds to distinct business workflows. Although promising, these approaches require the deployment of a large volume of LLMs as the number of IMAs increases, and hence are costly for edge devices. To this end, we aim to develop a novel paradigm that accomplishes various IMAs using a single compositional LLM. Achieving this goal involves two key challenges:

\noindent
\textbf{1) Guiding a single LLM to adapt to diverse IMA objectives:}
It is challenging for a learnable LoRA module in LLMs to simultaneously adapt to various tasks, as training parameters for one task can impair the performance of others. 

\noindent
\textbf{2) Ensuring the flexibility and efficiency of LLMs in resource-constrained mobile environments:}
LLMs for interactive multimodal applications tend to be deployed on edge devices, as cloud services suffer from several issues~\cite{ alghamdi2021data}, such as unreliable connectivity, high latency, and privacy concerns~\cite{wang1998consumer,iordache2021towards}. Due to the limited computational and memory resources of mobile devices, training compositional LLMs that are both flexible and efficient remains a non-trivial challenge for IMAs.

\subsection{Our Method}

The aforementioned issues motivate us to propose ContextLoRA and ContextGear. ContextLoRA is a novel method that incorporates multi-task workflows and their interrelationships into the fine-tuning process through a specifically designed training procedure, exploring the application of multimodal LLMs in interactive communication scenarios. We outline six high-level design principles for ContextLoRA and ContextGear to tackle the two challenges.

\noindent
\textbf{1) Flexible:}
Sub-models of ContextLoRA can be deployed independently or collaboratively as components of an LLM, which can be partitioned or integrated dynamically to adapt to different scenarios and requirements. 

\noindent
\textbf{2) Collaborative:}
Distributed edge devices can collaboratively fine-tune sub-models while executing complex processes and multiple tasks. 

\noindent
\textbf{3) Interpretable:}
Each sub-model corresponds to a specific sub-task. During training, logical relationships between sub-tasks are embedded into the model parameters, which enhances interpretability.

\noindent
\textbf{4) Pluggable:}
Sub-models can be dynamically added or removed, offering excellent scalability and simplified maintenance and updates.

\noindent
\textbf{5) Robust:}
The modular characteristics enhance the robustness of the model. When a portion of the data is compromised or subjected to attacks, it can be isolated and repaired, which limits the overall impact on the model.

\noindent
\textbf{6) Efficient:}
The algorithm is expected to be efficient, conserving both communication and computational costs to facilitate deployment on resource-constrained devices.

Keeping the above goals in mind, we first introduce ContextLoRA, a novel framework that endows an LLM with the ability to capture richly structured contextual information across IMAs via a task dependency graph. To support flexible model composition, we partition the learnable parameter matrix into distinct sub-matrices corresponding to each sub-task of IMAs. We then introduce a systematic fine-tuning method that comprises training, freezing, and masking phases and enables the model to capture latent inter-task dependencies. Therefore, our ContextLoRA enables an LLM to be dynamic, collaborative, interpretable, pluggable, and robust. Furthermore, we propose ContextGear to reduce computational and communication costs through a strategic grouping mechanism. Thus, it enables an LLM of IMAs to be efficient. Experiments on three benchmarks with 12 tasks show the effectiveness of our method. Our code is publicly available\footnote{https://github.com/caoxinye/ContextLoRA-and-ContextGear}.

\subsection{Main Contributions}
The main contributions of this paper are three-fold:
\begin{itemize}
    \item{\textbf{Compositional LLM with structured task relations}} We present ContextLoRA, a novel method that guides a multimodal LLM to learn to capture the rich structured context among IMAs by constructing a task dependency graph. We segment the LoRA parameter matrices for various tasks to facilitate a compositional LLM that is flexible, interpretable, robust, and pluggable, thus properly tackling the first challenge. To the best of our knowledge, we are the first to explore task relationships for the LoRA training parameters of interactive multimodal applications.
    
    \item{\textbf{Schedule strategy for collaborative fine-tuning:}} We present a hybrid pipeline parallelism for the collaborative fine-tuning of LLMs, comprehensively optimizing the computational and communication latency of interactive applications over wireless networks. The schedule strategy enables our proposed ContextGear to be flexible and efficient, thus tackling the second challenge. 
    
    \item{\textbf{Extensive experiments:}} We conduct extensive experiments on three benchmarks to show the effectiveness of the proposed ContextLoRA and ContextGear. We also deploy our method in a real-world environment that consists of three Jetson platforms. Finally, we provide a case study that visually demonstrates the detailed work procedure of the proposed ContextLoRA and give some insightful discussions based on our observations.
\end{itemize}

\subsection{Related Work}
\subsubsection{Multi-Task Fine-Tuning}
Current multi-task fine-tuning approaches are primarily based on mixture of experts (MoE) architectures~\cite{masoudnia2014mixture}, which introduces multiple expert models to handle different types of input. Existing research~\cite{lin2024teamlora,10.1145/3637528.3671609} emphasizes the use of MoE to coordinate existing LoRA experts without the need to specifically train expert parameters. Among various integration methods~\cite{salamancaseeded, muqeeth2024learning, zhao2024loraretriever}, the most common approach involves deploying a router to manage the interactions between these specialized models~\cite{ostapenko2024towards, huang2023lorahub}. However, existing MoE frameworks require complex interactions and long inference times. To improve reasoning capabilities and inference speed, we designed ContextLoRA. The two key differences between our work and previous works are: 1) ContextLoRA embeds multi-task relationships directly into model parameters, enhancing the model's understanding of the latent dependencies. 2) ContextLoRA enables compositional LLMs to be deployed on different devices independently without inter-task routing, thus achieving rapid inference.

\subsubsection{Parallelism for Edge Intelligence}
Parallelism has emerged as a crucial approach to accelerate the training of LLMs under resource-constrained scenarios~\cite{ryabinin2023swarm}. Contemporary parallel algorithms can be categorized into data parallelism and model parallelism. Specifically, data parallelism~\cite{VLDB20Torch, TDB15Petuum} partitions along the sample dimension to separate devices. Towards model parallelism, tensor parallelism~\cite{icpp22tesseract} splits a tensor across devices holding a subset of parameters. Meanwhile, pipeline parallelism~\cite{NEURIPS19GPipe, MLSYS21PipeMare, ICML21TeraPipe} is a model parallelism method that partitions the model by layers across devices and processes data in micro-batches sequentially through these partitions. Other research has explored hybrid parallelism approaches~\cite{hybrid1,hybrid2, SOSP19PipeDream}, which enable them to handle more specialized and complex scenarios. However, previous parallelism and optimization techniques are not specifically designed for multi-task fine-tuning, rendering them unsuitable for ContextLoRA. To address this limitation, we develop a pipeline parallelism strategy tailored for ContextLoRA and conduct joint optimization to accommodate the multi-task scenario effectively.

\begin{table}[!t]
\caption{DEFINITIONS OF NOTATIONS\label{tab:table1}}
\centering
\begin{tabular}{|c|c|}
\hline
 Notations  & Definitions\\
 \hline
        $G$ & Multi-task directed graph\\
         \hline
        $A$ & Adjacency matrix of $G$\\
         \hline
        $V$ & Set of task $v_i$\\
         \hline     
        $E$ & Set of directed edges of $G$\\
         \hline
        $W$ & LoRA matrix parameter\\
         \hline
        $S_k$ & Set of tasks with zero in-degree in k-th iteration\\
         \hline
        $\textbf{S}$ & Ordered list of $S_k$\\
         \hline
        $V_k$ & Set of tasks in k-th iteration\\
         \hline
        $P(v_j)$ & Prerequisite tasks of $v_j$\\
         \hline
        $\delta$ & Frozen ratio of matrix\\
        \hline
        $T_t$, $T_f$& Tasks in training and frozen state\\
         \hline
        $D_n$ & Set of all devices\\
        \hline
        $D_t$, $D_f$ & Set of devices for training and frozen matrix\\
         \hline
        $V_n$ & Set of tasks in pipeline\\
         \hline
        $V_t$, $V_f$ & Set of tasks allocated for $D_t$ and $D_f$\\
        \hline
        $G_c$ & Computation gap\\
        \hline
        $k$ & Batch size setup\\
         \hline
        $Q_n$ & Set of all possible model partitioning\\
         \hline
        $C$ & Total time consumption\\
         \hline
    \end{tabular}
    \label{tab:parameters}
\end{table}

\subsection{Paper Organization and Notations}

The remainder of the paper is organized as follows. Section \ref{2overview} presents the architecture of the proposed ContextLoRA and ContextGear. Section \ref{3ContextLoRA} describes our proposed ContextLoRA for multi-task reasoning. Section \ref{4ContextGear} presents our proposed ContextGear with pipeline parallelism and optimization. Section \ref{5dataset} demonstrates the construction process of the interactive multimodal dataset. Section \ref{6Exp} shows the experimental settings and discusses results compared with different baselines. Section \ref{7Dis} gives some insightful discussions. Finally, conclusions are drawn in Section \ref{8Conclusion}. Table \ref{tab:table1} lists the notations used in this paper.

\begin{figure*}[!t]
\centering
\includegraphics[width=1\textwidth,trim=30 31 25 10,clip]{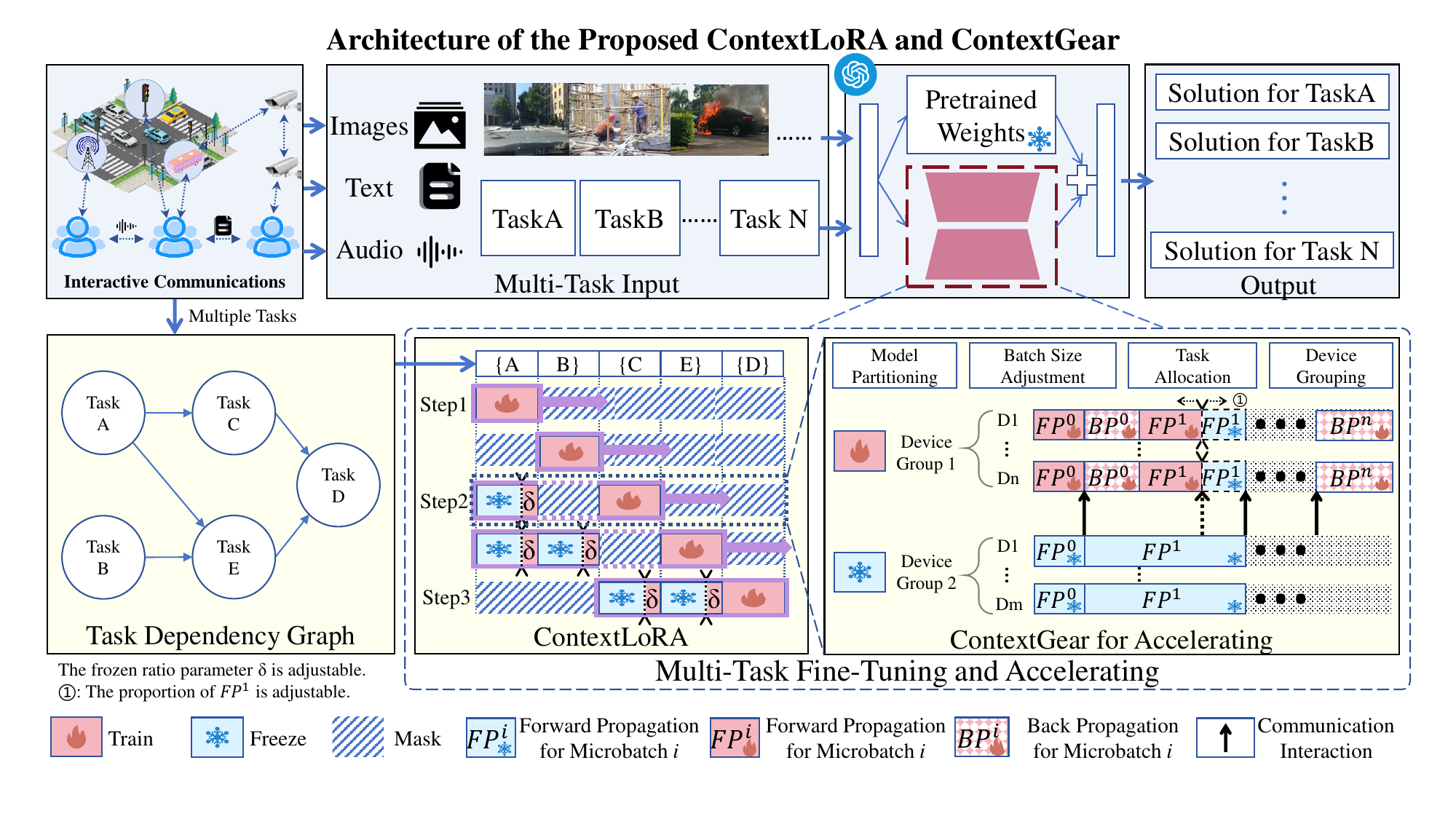}
\caption{Illustration of the proposed ContextLoRA and ContextGear architecture. In interactive communication scenarios, image and text inputs are fed into a multimodal LLM, which processes these inputs to generate solutions for multiple tasks. The ContextLoRA algorithm is proposed to partition and train the LoRA weight matrix based on task dependencies. In ContextGear, we design a pipeline for ContextLoRA to accelerate and optimize the training process.}
\label{fig:2}
\vspace{-4mm}
\end{figure*}

\section{System Overview}
\label{2overview}

To facilitate interactive multimodal communications, we present a collaborative multi-task fine-tuning and acceleration system, including the proposed ContextLoRA and ContextGear. As illustrated in Figure \ref{fig:2}, the system consists of multimodal inputs, a multimodal LLM, a multi-task fine-tuning method ContextLoRA, and an acceleration method ContextGear. Specifically, we construct a multimodal dataset, including the Internet of Vehicles, intelligent factories, and smart city scenarios with multiple tasks that encompass interactions among humans and machines. Images captured by sensors from the environment, along with users' textual or audio tasks, serve as inputs for fine-tuning the multimodal LLM. We hope the model can learn logical relationships among multiple tasks, which leads to the design of ContextLoRA. To accelerate the training process of ContextLoRA among edge devices, we propose ContextGear, a hybrid pipeline optimization algorithm for multi-task interactions.

\subsection{Traditional LoRA Fine-Tuning}

LoRA~\cite{hulora} is among the most widely used fine-tuning methods for large pre-trained models, known for its efficiency and scalability. Unlike traditional approaches that update all model parameters, LoRA introduces a novel technique to decompose the original weight matrix into two smaller, low-rank matrices. Specifically, during training, a learnable low-rank decomposition is applied, where matrices $A$ and $B$ modify the original weight matrix $W_0$. The resulting weight matrix is expressed as $W = W_0 + B A$, with $W_0 \in \mathbb{R}^{d \times k}$ remaining frozen. In this setup, $B \in \mathbb{R}^{d \times r}$ and $A \in \mathbb{R}^{r \times k}$, where $r$ represents the rank of the decomposition. By restricting updates to the low-rank matrices $A$ and $B$, LoRA greatly reduces trainable parameters and computational costs, as $r$ is typically much smaller than the dimensions of $W_0$.

\subsection{ContextLoRA}

The proposed ContextLoRA is a LoRA-based fine-tuning method for multi-task interactive scenarios. It first constructs a task graph based on multi-task relationships. Then, task dependencies are extracted from the multi-task graph into an ordered list. The LoRA weight matrix is correspondingly partitioned into multiple sub-matrices for each task. The core principle of ContextLoRA is to apply freezing, training, and masking operations to the submatrices based on task relationships. We use a sliding window to illustrate the sequential process of freezing and training the sub-matrices.

Our proposed ContextLoRA brings several benefits to traditional LoRA. First, unlike conventional LoRA and MoE architectures, ContextLoRA embeds task relationships into the weight matrix through a uniquely designed training process, enhancing the model's ability to understand multi-task dependencies. Second, since task relationships are embedded into the parameters, ContextLoRA eliminates the need for complex task-level interactions that are typical in MoE, enabling a more rapid response to multiple tasks. Lastly, ContextLoRA inherits the compositional ability of LoRA, enabling the deployment of different sub-tasks of a complex system among multiple devices for collaborative processing or combining parameters to form a unified model. Furthermore, it can even alter the model's emphasis on specific tasks by adjusting the frozen ratio of parameters.

\subsection{ContextGear}

To further advance the deployment of ContextLoRA on edge devices, we design a pipeline parallelism mechanism and a comprehensive optimization algorithm. Considering that frozen parameter matrices, unlike trainable matrices, do not require backward propagation, we develop a pipeline that divides devices into two groups: one group handles forward propagation of frozen parameters, while the other manages both forward and backward propagation of trainable parameters. Given that a task may correspond to multiple preceding tasks, meaning it could involve multiple frozen parameter matrices, we introduce a task allocation phase that allows certain frozen tasks to be reassigned to the trainable parameter device group. To balance pipeline workload, we introduce dynamic device grouping and task allocation in the pipeline design. Finally, we aim to minimize overall training time by adjusting the model partition point and batch size.

\section{ContextLoRA for Inter-Task Reasoning}
\label{3ContextLoRA}

\begin{figure}[!t]
\centering
\includegraphics[width=3.4in,trim=150 75 280 90,clip]{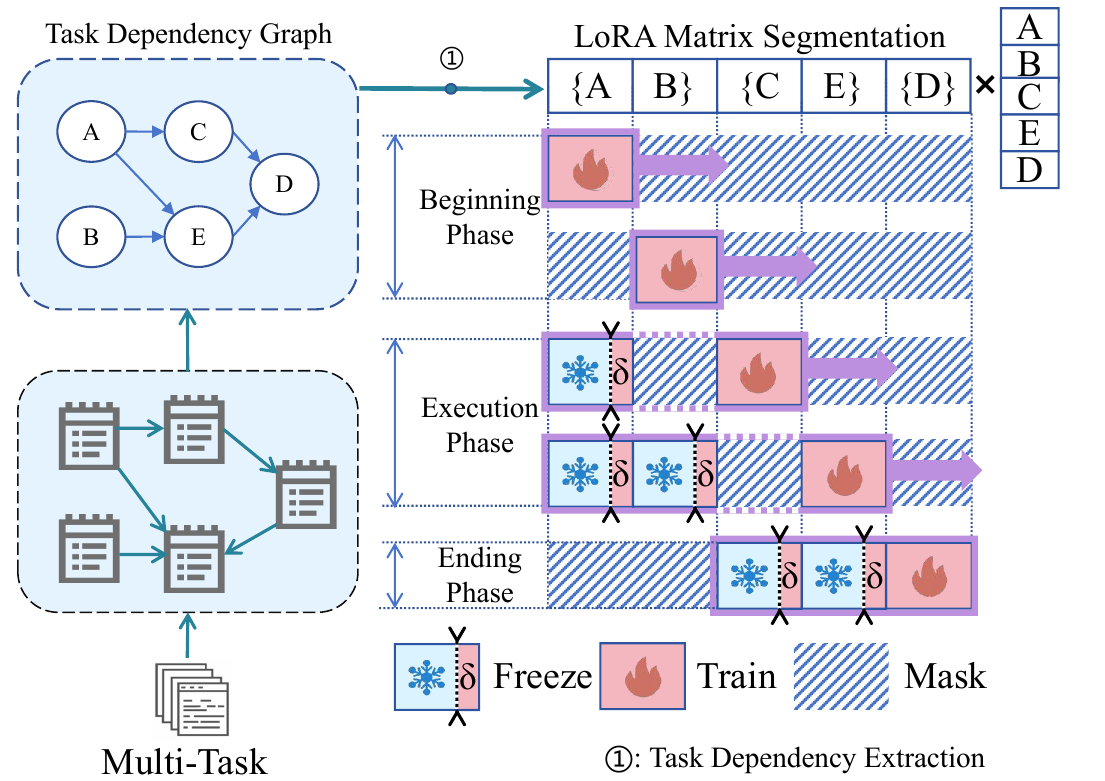}
\caption{Illustration of the proposed ContextLoRA. A task dependency graph is constructed to depict the relationship of multiple tasks. We apply ``freeze'', ``train'' and ``mask'' to map the task dependency into the parameter matrix.}
\label{fig_tocLoRA}
\end{figure}

The proposed ContextLoRA is a LoRA-based fine-tuning method for multi-task interactive scenarios. Specifically, we construct a directed graph to describe inter-task relationships. By setting the submatrices to three states: freeze, train, and mask, we embed task dependencies into the model parameters, as illustrated in Figure \ref{fig_tocLoRA}. Parameter $\delta$ is designed to adjust the frozen ratio, which is related to the focus between tasks. Generally, the algorithm can be divided into two main phases: graph-based task dependency extraction and sliding-window-based training of the partitioned LoRA matrix.

\begin{algorithm}[htbp]
\caption{ContextLoRA}
\label{a2}
\begin{algorithmic}
\STATE \textbf{Input:} Adjacency matrix $A$ generated by $G$
\STATE \textbf{Output:} Trained matrices $W = [W_1, W_2, \ldots, W_n]$

\FOR{$k \gets 1$ to L}
    \STATE $S_k \gets \emptyset$
    \FOR{$i \gets 1$ to $|V_k|$}
        \IF{$\sum_{j} A(V_k)[j][i] == 0$}
            \STATE $S_k \leftarrow S_k \cup \{v_i\} $
        \ENDIF
    \ENDFOR
    \STATE Append $S_k$ to $\textbf{S}$
    \STATE $V_{k+1} \gets V_k - S_k$
\ENDFOR
\FOR{$i \gets 1$ to L}
    \FOR{$v_j \in S_i$}
        \STATE $\mathcal{W}_{Train} \gets W[v_j] \cup W[P(v_j)](\delta)$
        \IF{$i==1$}
            \STATE $\mathcal{W}_{Freeze} \gets \emptyset$
        \ELSE
            \STATE $\mathcal{W}_{Freeze} \gets  W[P(v_j)](1-\delta)$
        \ENDIF
        \STATE ${W}_{Mask} \gets W - {W}_{Train} - {W}_{Freeze}$
        \STATE $W_{LoRA} \gets {W}_{Train} \cup {W}_{Freeze}$
        \STATE update $W[v_j]$ after LoRA fine-tuning $W_{LoRA}$
    \ENDFOR
\ENDFOR
\RETURN Trained matrices $W$
\end{algorithmic}
\end{algorithm}

\subsection{Graph-based Task Dependency Extraction}

The task dependency graphs are manually constructed based on the workflow of IMAs, which can be modified according to user commands or a company's pre-designed workflows. We represent the multiple tasks as a directed graph, where edges denote the dependencies between tasks. Subsequently, we convert the multi-task directed graph into an ordered list. The ordered list, which reflects the task execution sequence, is prepared for the upcoming fine-tuning of ContextLoRA.

The task dependency is represented by a directed graph $G = (V, E)$, where $V = \{v_1, v_2, \ldots, v_n\}$ represents the set of tasks, with each node $v_i$ corresponding to a specific task. $E \subseteq V \times V$ represents the directed edges between tasks, where an edge $(v_i, v_j) \in E$ indicates that task $v_i$ points to task $v_j$ meaning task $v_i$ must be completed before task $v_j$.
The adjacency matrix $A(V)$ is an $n\times n$ matrix that represents the directed graph $G$, where each element $A(V)[i][j]$ takes a value of either 0 or 1, can be defined as 
\begin{equation}
A(V)[i][j] = 
\begin{cases} 
1,&(v_i, v_j) \in E \\
0,&(v_i, v_j) \notin E
\end{cases}
\ i,j \in\left[1,n\right].
\end{equation}

In the task dependency extraction process, tasks are sorted based on their relationships within a task dependency graph. Nodes with zero in-degree are called source nodes, meaning that no other tasks depend on them, so they do not require prior task completion. Conversely, nodes with zero out-degree are called sink nodes, as they do not have outgoing edges, implying that these tasks do not influence any subsequent tasks. This extraction process iteratively identifies the source nodes of the task graph, inserts them into the list $\textbf{S}$, and deletes them from the graph. As shown in Algorithm \ref{a2}, the process repeats until all nodes are removed from task set $V$ after $L$ times of iteration. In the k-th iteration, the set of source nodes of the current task set $V_k$ is denoted as $S_k$. The ordered list composed of all $S_k$ is denoted as $\textbf{S}$. Additionally, we define $V_1 = V$. The procedure can be expressed as
\begin{equation}
\textbf{S} = [S_1, S_2, \cdots, S_L], 
\end{equation}
\begin{equation}
S_k = \{v_j \in V_k \mid \forall i, A(V_k)[i][j] = 0 \}, k \in\left[1, L\right], 
\end{equation}
\begin{equation}
V_{k+1} = V_k - S_k.
\end{equation}

\subsection{Sliding-window Training of the Partitioned LoRA Matrix}
The training process involves iterative training of the LoRA parameter matrices $W = [W_1, W_2, \ldots, W_n]$, which are obtained by splitting the overall LoRA matrix $W$ into $n$ column-wise segments. Each segment $W_i$ corresponds to a specific task $v_i$ in the ordered list $S$. As shown in Algorithm \ref{a2}, we traverse all the $S_i$ in $\textbf{S}$. For every task $v_j$ in $S_i$, we partition the parameter matrix W into three distinct blocks: trainable parameters, frozen parameters, and masked parameters. Following this decomposition, we isolate the masked parameter block and perform LoRA fine-tuning on the remaining parameters $W_{LoRA}$. During this process, gradient updates are exclusively applied to the trainable parameter block while maintaining the frozen parameters in a static state. For each task $v_j$, only the training parameters $W[v_j]$ are updated in W. Sliding-window-based training process is summarized as follows:
(1) For the initial task set $S_1$: we train the parameter matrix $W[v_j]$ for each task $v_j \in S_1$, while keeping other parameter matrices masked. Since no task in $S_1$ has prerequisites, no parameter matrices are frozen in this phase.
(2) For subsequent task set $S_i (i\in [2,L])$: we train the parameter matrix $W[v_j]$ for each task $v_j \in S_i$, while freezing the parameter matrices $W[P(v_j)]$ corresponding to its prerequisite tasks $P(v_j)$, where $P(v_j)$ is defined as 
\begin{equation}
P(v_j) = \{v_i \in V \mid A(V)[i][j] = 1\}.
\end{equation}
$W(\delta)$ denotes that there are $\delta$\% columns of parameters that are activated. During the training of each task $v_j \in S_i$, the status of the parameter matrices can be categorized as
\begin{equation}
\begin{aligned}
&{W}_{Freeze} = W[P(v_j)](1-\delta),\\
&{W}_{Train} = W[v_j] \cup W[P(v_j)](\delta),\\
&{W}_{Mask} = W-{W}_{Train}-{W}_{Freeze}.\\
\end{aligned}
\end{equation}

It is worth noting that ${W}_{Train}$ refers to LoRA fine-tuning of the parameter matrix, while ${W}_{Freeze}$ means that the parameters are not updated but can still serve as features for the training module. ${W}_{Mask}$ indicates that the corresponding parameters do not participate in any training process. When applying the proposed ContextLoRA method, each task-specific parameter matrix undergoes fine-tuning, and only the task matrices from the previous step are involved in their respective training processes, thereby achieving the mapping from task relationships to fine-tuned parameters.

To provide a clearer illustration of the proposed ContextLoRA workflow, we instantiate a task graph, illustrated in Figure \ref{fig_tocLoRA}, that can be divided into three phases.
(1) Beginning Phase: Train the parameter matrices corresponding to tasks $A$ and $B$. At this stage, all other parameter matrices are masked and do not participate in the training process.
(2) Execution Phase: First, train the parameter matrix corresponding to task C. The parameter matrix for task A, which serves as a prerequisite for task C, is frozen at a ratio of $1-\delta$. All other parameter matrices are masked. Next, train the parameter matrix for task $E$ with the parameter matrices for tasks $A$ and $B$ frozen, and all remaining parameter matrices masked.
(3) Ending Phase: Finally, freeze the parameter matrices for tasks $C$ and $E$, keep all other matrices masked, and train the parameter matrix corresponding to task $D$.

We introduce a configurable parameter $\delta$ to adjust the frozen ratio of the LoRA matrix. By varying the frozen ratio, we can control the model's focus. A frozen ratio of 100\% implies that subsequent tasks do not influence prior tasks, resulting in better independence of each sub-model. As the frozen ratio decreases, the influence of subsequent tasks on prior tasks increases, and the model places more emphasis on the final outcome, especially the output at the last node. In other words, if we desire a model capable of both independent operation and collaborative completion of complex tasks, we can increase the frozen ratio. Conversely, if our primary interest lies in the final output of the model, and sub-models are merely intermediate processes without a need for independent outputs, we can reduce the frozen ratio.

Our directed graph includes a combination of parallel and serial processes but does not incorporate any cyclic structures. This omission is deliberate, as cyclic structures represent iterative updates of tasks, which leads to repeated training of the associated parameter matrices without meaningful benefit. Instead, such iterative processes can be handled through multiple rounds of LLM dialogue.

\section{The proposed ContextGear}
\label{4ContextGear}
To prompt LLM fine-tuning on edge devices, we design pipeline parallelism as the scheduling strategy and an optimization algorithm for training acceleration. The pipeline is designed to accommodate the unique training method of ContextLoRA, assigning the two device groups, respectively, to the ``Train'' and ``Freeze'' parameter matrices within ContextLoRA. The optimization algorithm aims to balance the workload and minimize the training time of ContextLoRA by focusing on four key aspects: model partitioning, batch size adjustment, task allocation, and device grouping.

\subsection{Pipeline Parallelism for the Proposed ContextLoRA}
\begin{figure}[!t]
\centering
\includegraphics[width=0.48\textwidth,trim=60 82 270 30,clip]{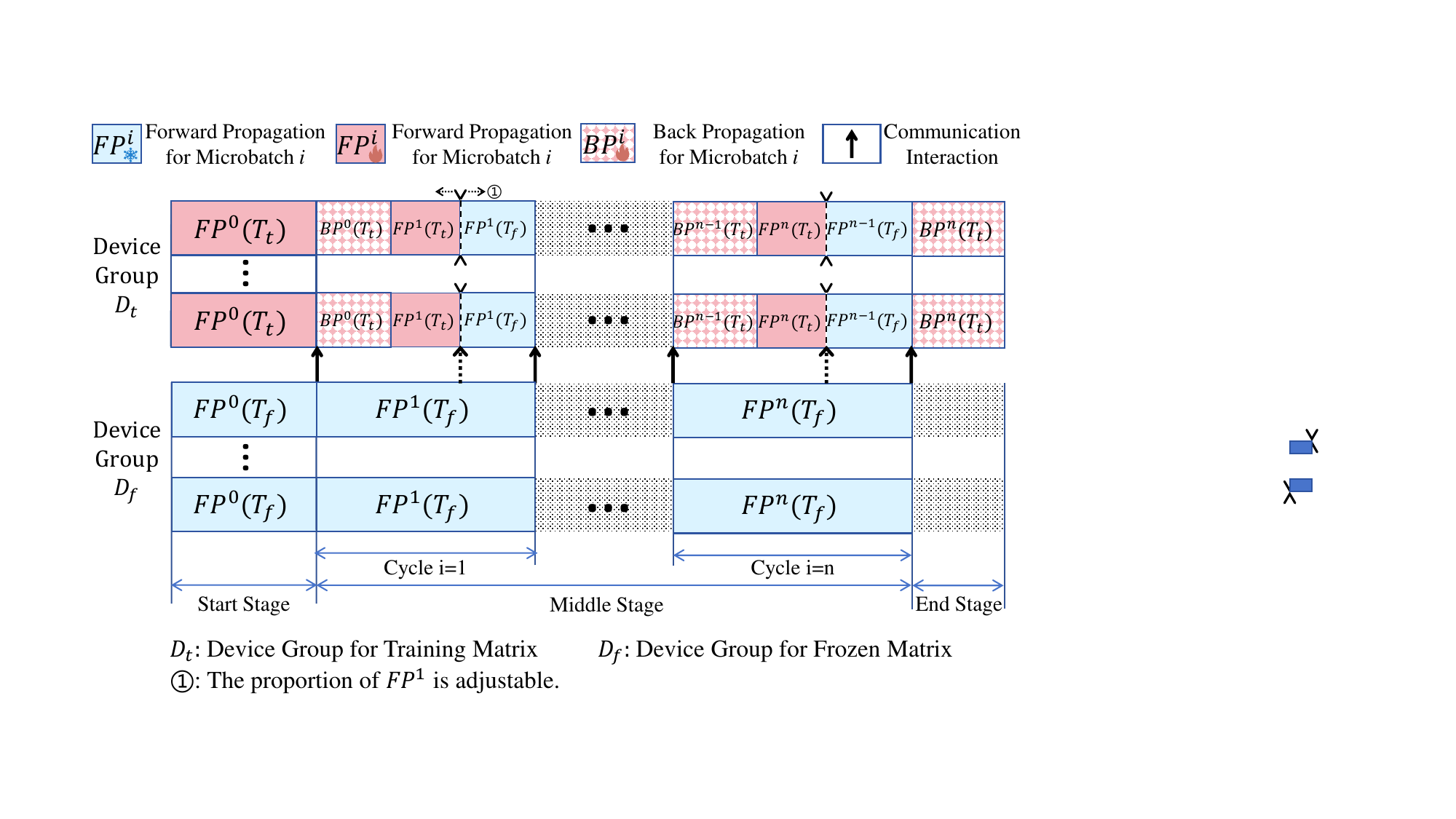}
\caption{Illustration of pipeline design of the proposed ContextGear. Devices are divided into two groups for the training and freezing operations. The device group for training $D_t$ involves model forward and backward propagation while the device group for frozen operation $D_f$ only involves forward propagation. To balance the task load, a part of the tasks in $D_f$ can be transmitted to $D_t$.}
\label{fig_PP}
%\vspace{-3mm}
\end{figure}

We design ContextGear, a novel pipeline parallelism approach for multi-task interactive scenarios, accommodating the freezing and training processes in the proposed ContextLoRA. We consider two sets of tasks: tasks in the frozen state, denoted as $T_f$, and tasks in the training state, denoted as $T_t$. The pipeline design is illustrated in Figure \ref{fig_PP}.

To formulate the pipeline process, let $FP^i(T_t)$ and $BP^i(T_t)$ represent the forward and backward propagation of $T_t$ for batch $i$ respectively. In the first stage, both $T_f$ and $T_t$ perform forward propagation of batch 0 on their respective devices, characterized as $FP^0(T_f)$ and $FP^0(T_t)$. In the middle stage, we design multiple iterative cycles in the pipeline to maximize the utilization of each device, thereby minimizing the overall training time. In each cycle $i$, the backward propagation of batch $i-1$ is performed solely for $T_t$, while the forward propagation of batch $i$ is conducted for both $T_f$ and $T_t$. Since the parameters associated with $T_f$ are frozen, there is no backward propagation involved in these tasks. In the last stage, $T_t$ executes backward propagation of the last batch n. 
The detailed operations in three stages of the designed pipeline can be outlined as follows:
\begin{equation}
\begin{aligned}
&\text{First stage:}\quad FP^0(T_f),\ FP^0(T_t), \\
&\text{Middle stage:}\quad FP^i(T_f),\ FP^i(T_t),\ BP^{i-1}(T_t),\\
&\text{Last stage:}\quad BP^{n}(T_t).
\end{aligned}
\end{equation}

It is also worth noting that each training task may correspond to the multiple frozen parameter matrices of prerequisite tasks involved in forward propagation. To balance the computational load across different device groups, we introduce a dashed line in Figure \ref{fig_PP}, which represents the manner in which forward propagation tasks $FP^1(T_f)$ associated with frozen parameters on device group $D_f$ can be offloaded to the device group $D_t$ responsible for training tasks. This ensures efficient load balancing across the pipeline stages.

The pipeline parallelism designed in our proposed ContextGear enables efficient utilization of computational resources by scheduling forward and backward propagation tasks across multiple devices, thereby reducing idle time. Furthermore, dividing the training process into small cycles ensures that both frozen and training tasks are performed in a pipeline manner.

\subsection{Optimization for the Proposed Pipeline Strategy}
To further enhance the efficiency of the pipeline strategy described above, we propose an optimization algorithm that minimizes the total training time by model partitioning, batch size, task allocation adjustment, and device grouping. This algorithm aims to balance computational and communication loads among devices, achieving minimum latency.

\begin{algorithm}[htbp]
\caption{Optimization for Pipeline Strategy}
\label{alg:opt}
\begin{algorithmic}
\STATE \textbf{Input:} $D_n$, $V_n$, $Q_n$, $k_{max}$, 
\STATE \textbf{Output:} $D_t, V_t, Q, k, C_{min}$
\STATE $C_{min} \gets \infty$
    \FOR{$V_t \in V_n$}
        \FOR{$D_t$ in $D_n$}
            \STATE ${G_c} \gets |R_c(D_t)-R_r(V_t)|+|R_c(D_f)-R_r(V_f)|$
            \STATE update $D_t, V_t$ to minimize $G_{c}$
        \ENDFOR    
    \ENDFOR
    \FOR{$Q \in Q_n$}
        \FOR{$k \gets 1$ to $k_{max}$}
            \STATE $C \gets \max \{ C_T(Q_t, k), C_F(Q_f, k)\}$
            \STATE update $Q, k$ to minimize $C$
            \STATE update $C_{min}$
    \ENDFOR
\ENDFOR
\RETURN $D_t, V_t, Q, k, C_{min}$
\end{algorithmic}
\end{algorithm}

\paragraph{Model Partitioning}
We dynamically adjust the model's partition to balance the computational load across devices within each device group. By optimally distributing the model layers among devices, we prevent any single device from becoming a bottleneck.

\paragraph{Batch Size Adjustment}
We adjust the batch size of the model to ensure that each device can process its assigned workload efficiently without exceeding its computational capacity.

\paragraph{Task Allocation}
Tasks are allocated to minimize communication overhead. We consider the available computational capacity of each device group to assign tasks effectively, reducing the data transmission time between devices.

\paragraph{Device Grouping}
As the proposed ContextLoRA is divided into two device groups to handle the training and frozen modules simultaneously, edge devices can be assigned to different groups based on computational capacity and task requirements.

These optimizations are essential for balancing computational and communication costs, ultimately improving the overall efficiency of pipeline parallelism in edge LLM applications. 
Overall, the optimization process focuses on two key aspects: pipeline design to balance computational resources and model partitioning with batch size adjustment to minimize training time.

To formulate the proposed ContextGear, we define the set of all devices as $D_n$, which is divided into the device group $D_t$ for the training matrix and the device group $D_f$ for frozen matrices, where $D_t \cup D_f = D_n$. \(R_c(D_t)\) and \(R_c(D_f)\) respectively represent the computational capacities of $D_t$ and $D_f$. $V_n$ denotes the set of all tasks divided into task sets $V_t$ and $V_f$ allocated to $D_t$ and $D_f$, where $V_t \cup V_f = V_n$. \(R_r(V_t)\) and \(R_r(V_f)\) respectively indicate the computational requirements of $V_t$ and $V_f$.

The disparity between device computational capacity and task computational requirements leads to the computational gap $G_{c}$. The total computational gap $G_{c}$ consists of the gaps in the training and frozen pipeline design. Minimizing this gap $G_{c}$ is essential to fully utilize the computational potential of edge devices, and it is expressed as follows:
\begin{equation}
G_{c} = \min_{D_t}\min_{V_t}(|R_c(D_t)-R_r(V_t)| + |R_c(D_f)-R_r(V_f)|).
\end{equation}
Task allocation procedure indicates that part of the forward propagation of frozen task can be done in device group $D_t$ adaptively.

As shown in Algorithm \ref{alg:opt}, we update $D_t, V_t$ to minimize $G_{c}$. After optimizing the computational gap, we identify the device groups and task allocations associated with the training and freezing matrix pipelines. Next, we optimize the overall training time on the pipeline from two perspectives: model partitioning and batch size adjustment. The overall time consumption \( C_{T} \) of the proposed ContextGear among edge device group $D_t$ can be expressed as:

\begin{align}
C_{T} = &\max ( \sum_{l=1}^{L} (T_{\text{fwd}}(l, D_t) + T_{\text{bwd}}(l, D_t))
\notag
\\& +\sum_{d_i, d_j \in D_t} T_{\text{comm}}(d_i \to d_j)).
\end{align}
The overall training time \( C_{F} \) for frozen model deployed on $D_f$ can be expressed as:
\begin{equation}
C_{F} = \max ( \sum_{l=1}^{L} (T_{\text{fwd}}(l, D_f)+\sum_{d_i, d_j \in D_f} T_{\text{comm}}(d_i \to d_j)), 
\end{equation}
where \( L \) is the total number of layers in the model.

\( T_{\text{fwd}}(l, D_n) \) and \( T_{\text{bwd}}(l, D_n) \) are the forward and backward propagation time, respectively, for layer \( l \) of the model.
\( T_{\text{comm}}(d_i \to d_j) \) is the communication time required to transmit data between devices \( d_i \) and \( d_j \). 
The communication interactive time \( T_{\text{comm}} \) can be influenced by many factors such as transmitted data size, the bandwidth, and the inherent latency in the communication process.

The ultimate time consumption of the pipeline is determined by the longer time cost between $C_T(Q_t,k)$ and $C_F(Q_f,k)$ corresponding to device group $D_t$ and $D_f$. The core of our approach lies in dynamically adjusting model partitioning and batch size to prevent any device from becoming a bottleneck. The total time consumption \( C \) of the pipeline is denoted as
\begin{align}
C = \min_{Q}\min_{k}\max\{C_T(Q_t,k),C_F(Q_f,k)\}, 
\end{align}
where $Q$ denotes the model partitioning across $D_n$ and $k$ is the batch size of the model. $Q_n = [q_1,\cdots,q_i,\cdots,q_N]$ is the set of all possible model partitioning and $Q \in Q_n$. $Q = Q_t \cup Q_f$. $N$ denotes the number of layers. $Q_t$ is the first \( N_t \) elements of \( Q_n \) correspond to device group \( D_t \), while $Q_f$ is the last $N_f = N - N_t$ elements correspond to device group \( D_f \). As shown in Algorithm \ref{alg:opt}, we update $Q, k$ to minimize $C$. Under varying computational and hardware constraints, the optimal model partitioning Q and batch size k can differ substantially. For each resource configuration, we perform a systematic exploration of Q and k to identify the combination that minimizes end-to-end runtime. The profiling overhead associated with this tuning phase is negligible relative to the overall training time.

\begin{figure}[!t]
\centering
\subfloat[Internet of Vehicles\ \ \ \ \ \ \ \ \ \ \ \ ]{\includegraphics[width=0.48\textwidth,trim=230 160 105 70,clip]{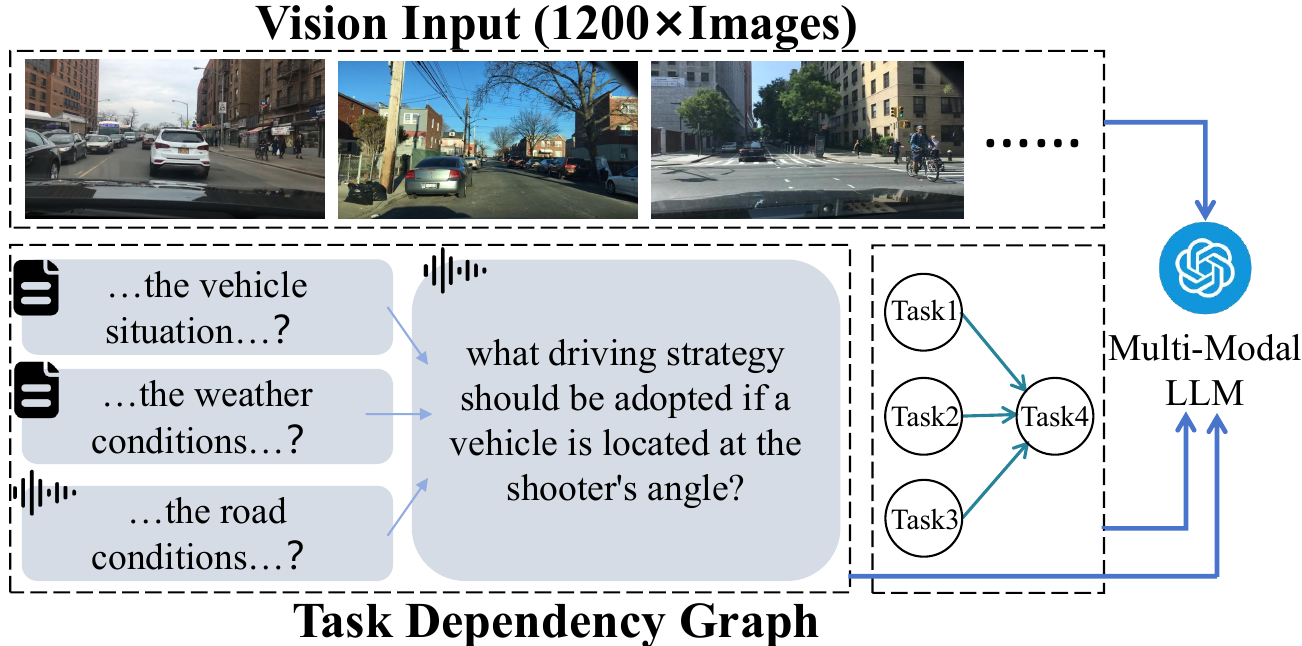}
\label{fig:traffic}}
\hfil
\subfloat[Intelligent Factory\ \ \ \ \ \ \ \ \ \ \ \ \ \ ]{\includegraphics[width=0.48\textwidth,trim=230 115 130 70,clip]{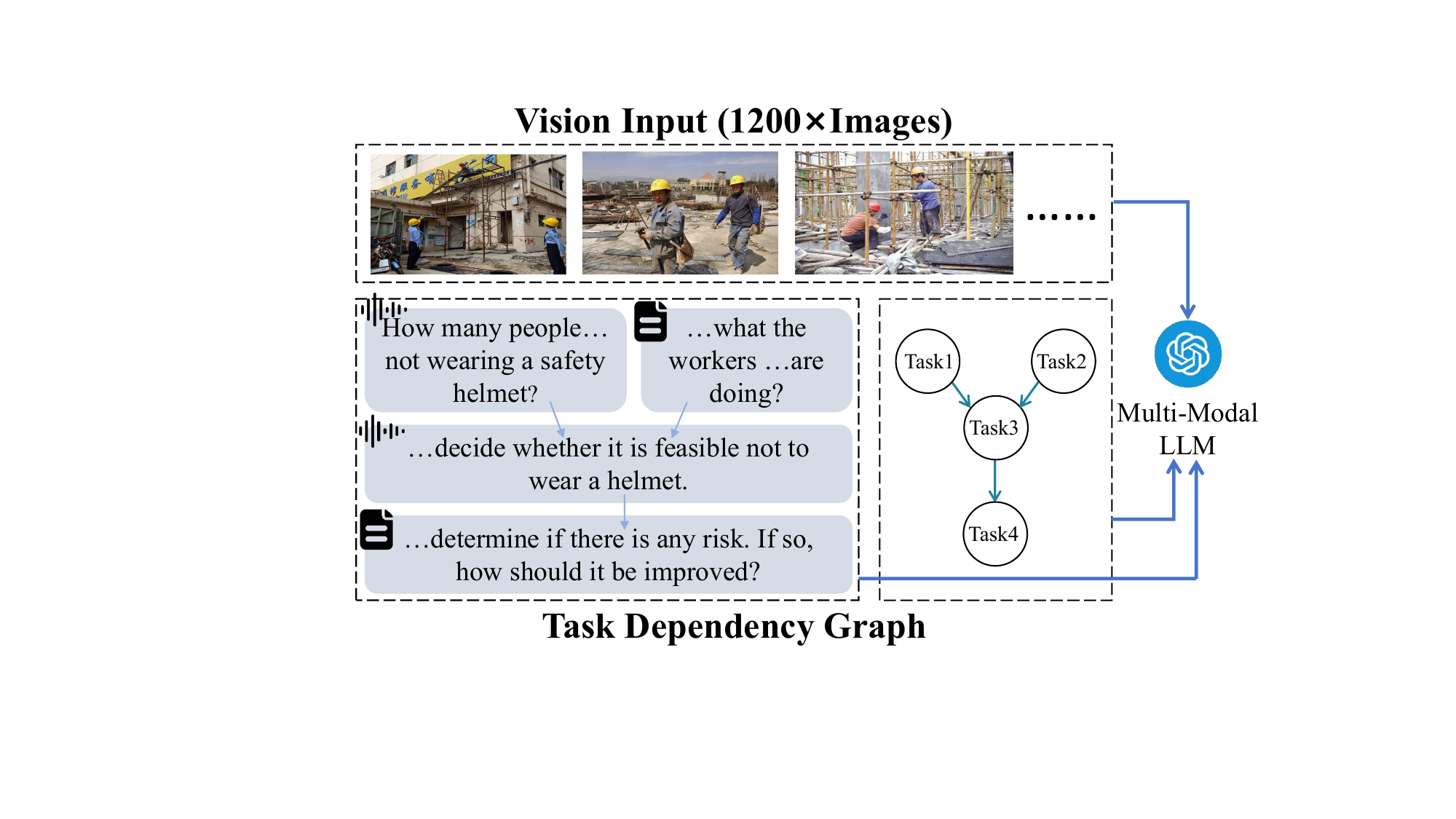}
\label{fig:helmet}}
\hfil
\subfloat[Smart City\ \ \ \ \ \ \ \ \ \ \ \ \ \ \ \ \ \ ]{\includegraphics[width=0.48\textwidth,trim=230 105 105 70,clip]{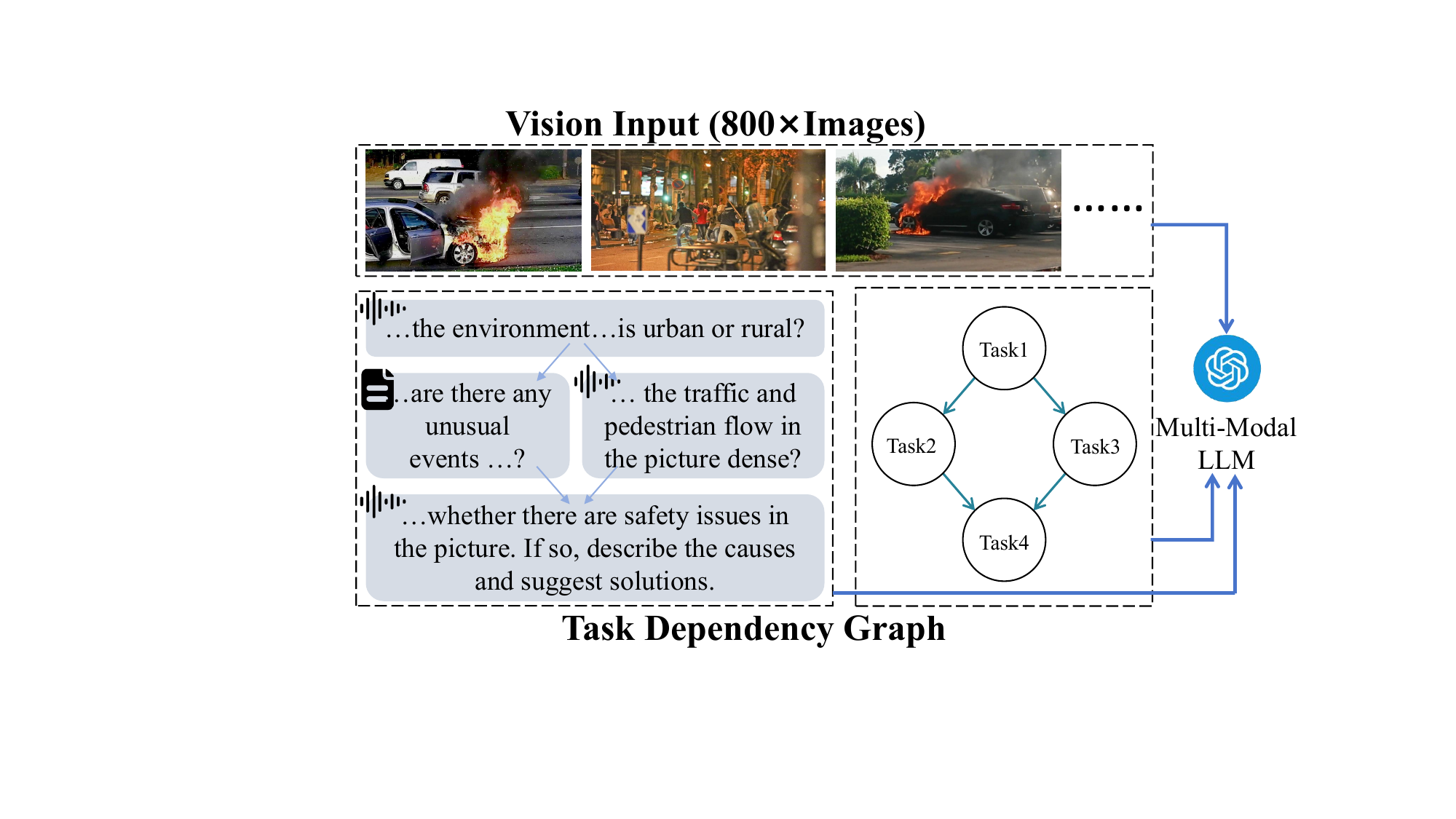}
\label{fig:RAD}}
\caption{Illustration of the constructed computer vision datasets with multiple tasks under three scenarios. The multimodal LLM inputs include images, task descriptions, and task graphs.}
\label{fig_questions}
\vspace{-3mm}
\end{figure}

\section{Computer Vision Dataset for Multi-Task Interactive Communications}
\label{5dataset}

We construct multi-task datasets related to interactive multimodal communications scenarios, including the Internet of Vehicles, intelligent factories, and smart cities. In each scenario, four Q\&A pairs are designed, which, together with the corresponding images, form the datasets for our model. Each dataset contains 1,200 images. These Q\&A pairs are structured to create distinct task dependency graphs that reflect the logical relationships between tasks. As a result, the data distribution across tasks is highly balanced. Labels are manually assigned, and the task dependency graphs are manually created based on the workflow of interactive multimodal applications. As illustrated in Figure \ref{fig_questions}, the tasks become progressively more complex and in-depth as they build upon the foundations established by the earlier ones.

For example, in the scenario of the Internet of Vehicles, we select a dataset of traffic record images\footnote{Traffic records: https://dl.cv.ethz.ch/bdd100k/data/}, and construct three basic questions including ``Describe the vehicle situation in the diagram", ``Describe the weather conditions in the picture.", and ``Describe the road conditions in the picture.". Based on these, we propose an open-ended question: ``Taking into account all the elements mentioned above, what driving strategy should be adopted if a vehicle is located at the shooter’s angle?". Similarly, for intelligent factories and smart city scenarios, we use an image dataset of safety helmets\footnote{Helmet: https://aistudio.baidu.com/datasetdetail/50329/0} and a custom dataset focusing on road abnormalities such as road accidents and fighting\footnote{RAD: https://github.com/sarfarazmemon/RAD/tree/main}, respectively.

\section{Experiments}
\label{6Exp}

\subsection{Experimental Setup}

We conduct the simulation on a server with Ubuntu 20.04.5 LTS operating system. The hardware configuration includes 64GB of RAM, an Intel(R) Xeon(R) Silver 4210 processor, and two NVIDIA A40 GPUs, each equipped with 48GB of memory. 
For real-world tests, we use three Jetson AGX Orin 64GB embedded devices. Each embedded device features a 2048-core NVIDIA Ampere GPU with 64 Tensor Cores, a 12-core Arm Cortex-A78AE CPU, and 64GB LPDDR5 memory. 

For the fine-tuning process, the training sets of the IoV and Intelligent Factory datasets each contain 4,800 Q\&A pairs, while the smart city dataset contains 3,200 Q\&A pairs. Each dataset's test set consists of 400 Q\&A pairs. We perform fine-tuning based on the LLaVA~\cite{NEURIPS2023_6dcf277e} model. The fine-tuning hyperparameter settings follow those of LLaVA, with a batch size of 4 and a total of 8 epochs. The hyperparameter settings for fine-tuning are presented in Table \ref{tab:hyperparams}.

\begin{table}[ht]
  \centering
  \caption{Training Hyperparameters}
  \label{tab:hyperparams}
  \begin{tabular}{|c|c|}
\hline
    \textbf{Hyperparameter}       & \textbf{Value}       \\ 
    \hline
    Batch Size           & 4            \\
    \hline
    lr                   & 2e-4         \\
    \hline
    lr Schedule          & cosine decay \\
    \hline
    lr Warmup Ratio      & 0.03         \\
    \hline
    Weight Decay         & 0            \\
    \hline
    Epoch                & 4            \\
    \hline
    Optimizer            & AdamW        \\
    \hline
    DeepSpeed Stage      & 3            \\
    \hline
  \end{tabular}
\end{table}

\subsection{Implementation Details}

\begin{figure*}[!t]
\centering
\includegraphics[width=1\textwidth,trim=29 115 45 160,clip]{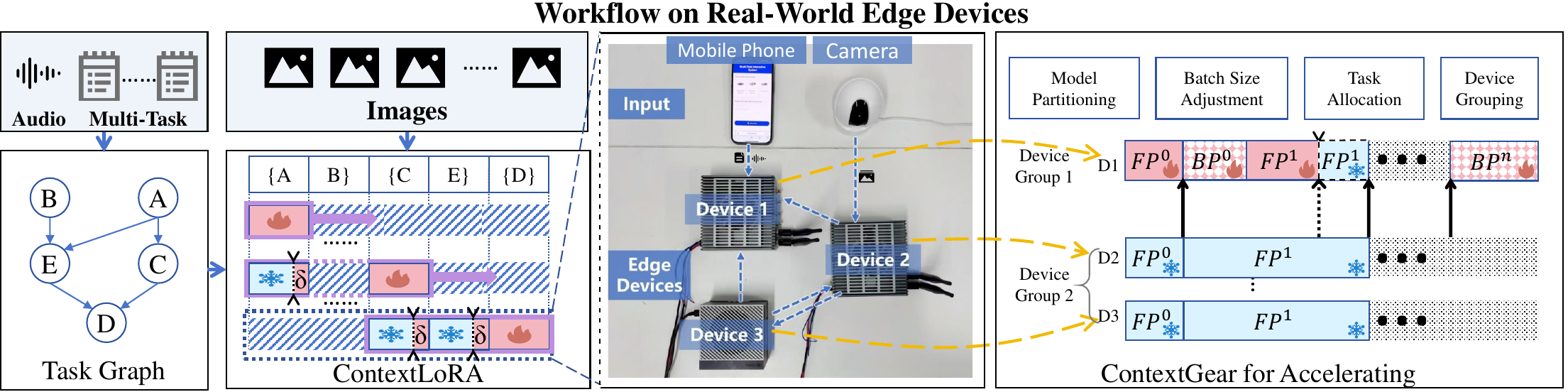}
\caption{Illustration of the workflow on real-world edge devices. Training of our ContextLoRA conducted among three edge devices can be accelerated by the proposed ContextGear. In this real-world testbed, we use a mobile phone and a camera to engage with the system using audio, text, and images.}
\label{fig_jetson}
\vspace{-5mm}
\end{figure*}

In Figure \ref{fig_jetson}, we illustrate the implementation of the ContextLoRA training process and the ContextGear pipeline acceleration on three edge devices. These devices are divided into two groups: one handles the forward and backward propagation of the ContextLoRA training matrix, while the other handles the forward propagation of the frozen matrix in ContextLoRA. In this real-world testbed, we use a mobile phone and a camera to engage the system with audio, text, and images.

During the training phase, multiple devices collaborate to accelerate the model fine-tuning process. In the inference phase, each device is responsible for specific task modules. For instance, in the Internet of Vehicles scenario, Device 1 utilizes cameras to analyze road conditions and identify traffic congestion and obstacles. Device 2 monitors weather conditions and Device 3 performs the final decision making. Task modules are designed for modular integration and can be dynamically adjusted based on application needs. For example, if a user interaction only requires road condition updates, activating Device 1 alone is sufficient. When users need full interaction with the model and comprehensive decision making, all devices are activated. This strategy simplifies inter-device communication while ensuring reliable support for intelligent applications in complex scenarios.

\subsection{Baselines}

We compare our proposed ContextLoRA with several state-of-art baselines, including \textbf{HydraLoRA} (an asymmetric LoRA architecture)~\cite{tian2024hydralora}, \textbf{Mixture of LoRA Experts}(MoLE)~\cite{wumixture}, and the original LoRA~\cite{hulora}. HydraLoRA and MoLE are two LoRA-based methods designed for fine-tuning multi-task modules. Original LoRA takes multi-task inputs to train a whole parameter matrix. Different from the above baselines, our proposed ContextLoRA partitions the LoRA matrix, enhancing the reasoning ability of LLM.

The proposed ContextGear is compared with several baselines, including \textbf{JoRA}~\cite{Acl24Jora} and \textbf{deepspeed}~\cite{arxiv2020ZeRO}. JoRA is a tensor-parallel LoRA library for fine-tuning, while DeepSpeed is an optimization library supporting tensor, model, and pipeline parallelism. Different from the above baselines, our proposed ContextGear introduces a hybrid parallelism to optimization model training in multi-task interactive communications.

\subsection{Metric}
 
We use the accuracy of multiple-choice questions as a metric to evaluate the inference capability of different methods. For each model's responses, a scoring system is applied. Specifically, if all selected options in the response perfectly match the correct answers, the model is awarded 1 point. If some but not all selected options match, the response receives 0.5 points. Responses with mismatched options or errors in the output format are assigned 0 points. The final accuracy between 0 and 1 is calculated as the total score divided by the maximum possible score.

\subsection{Performance}

\begin{figure*}[!t]
\centering
\subfloat[]{\includegraphics[width=2.2in]{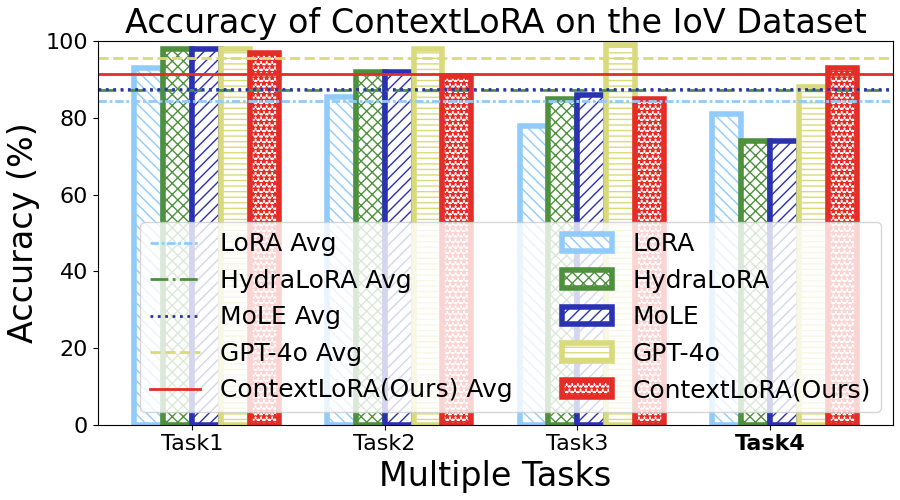}
\label{fig:7_first_case}}
\hfil
\subfloat[]{\includegraphics[width=2.2in]{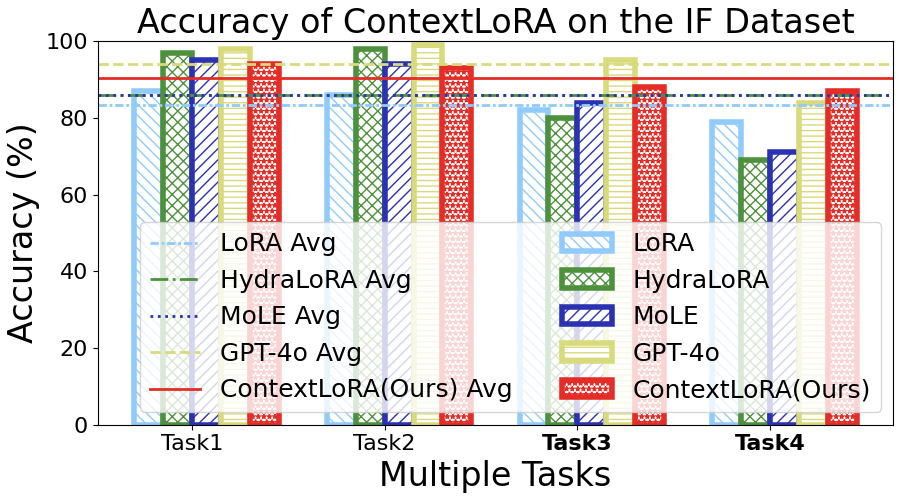}
\label{fig:7_second_case}}
\hfil
\subfloat[]{\includegraphics[width=2.2in]{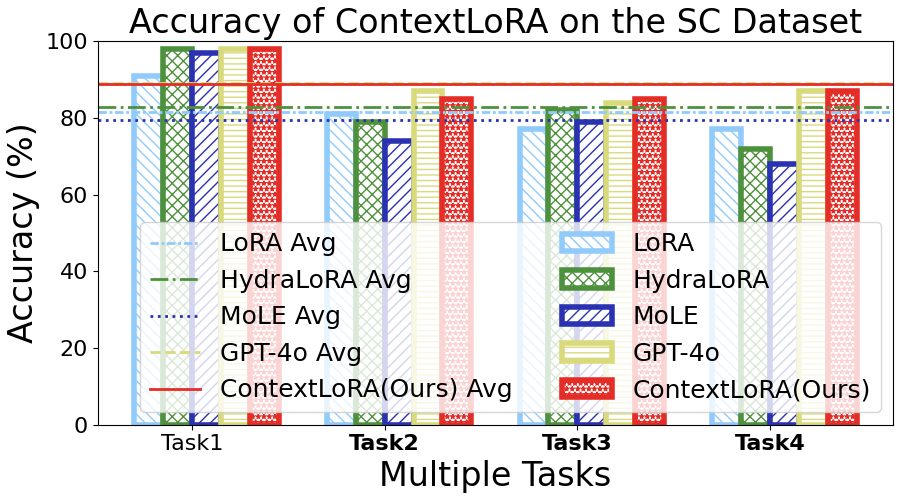}
\label{fig:7_third_case}}
\hfil
\subfloat[]{\includegraphics[width=2.2in]{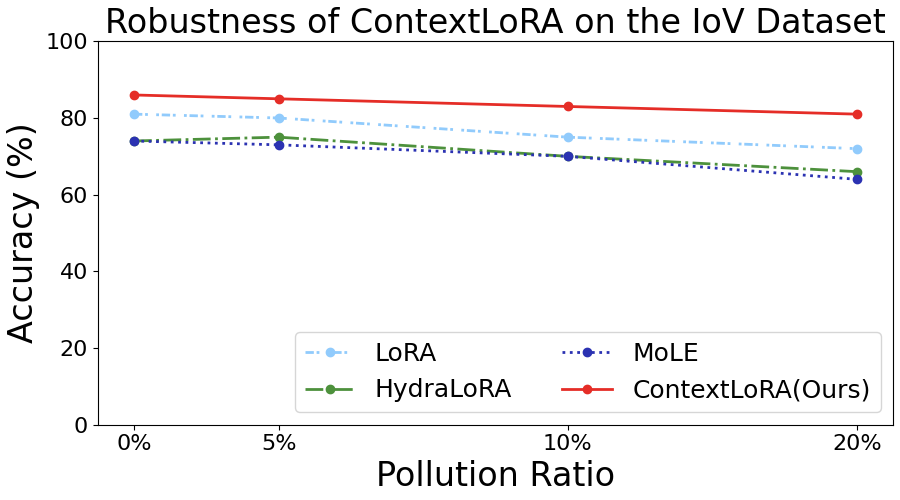}
\label{fig:7_fourth_case}}
\hfil
\subfloat[]{\includegraphics[width=2.2in]{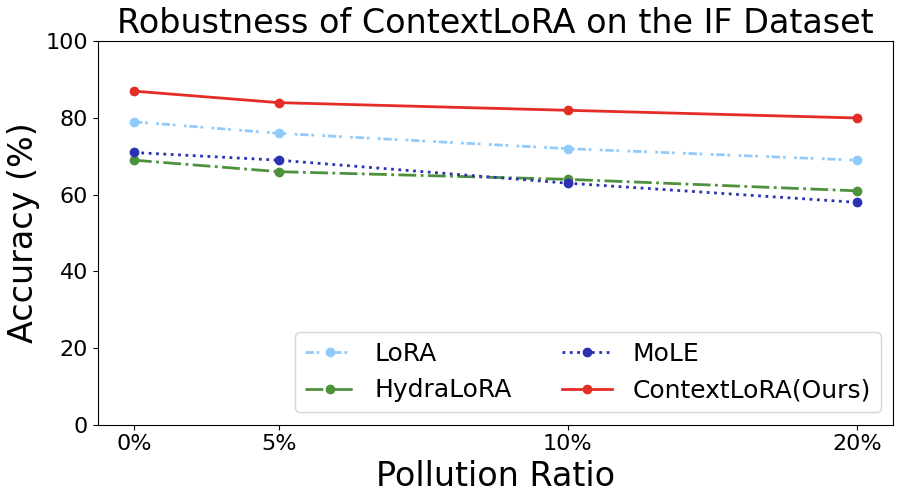}
\label{fig:7_fifth_case}}
\hfil
\subfloat[]{\includegraphics[width=2.2in]{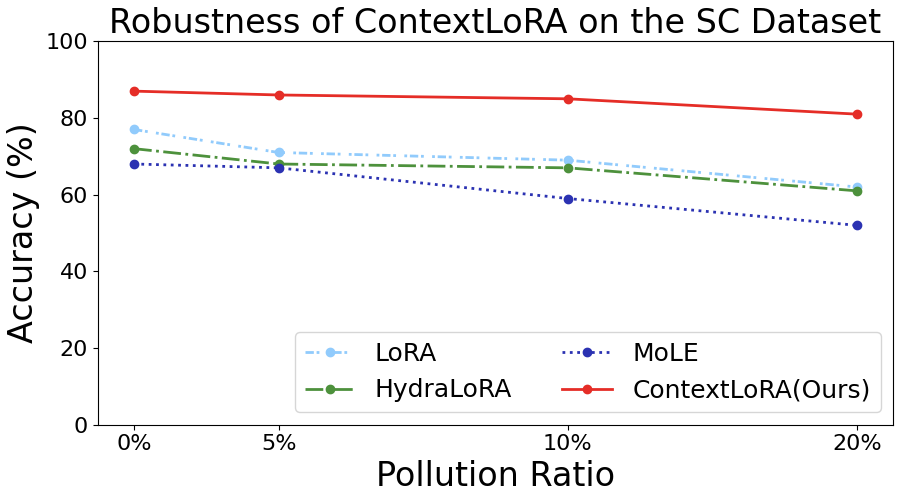}
\label{fig:7_sixth_case}}

\caption{Performance comparisons on three datasets. (a), (b), and (c) compare the accuracy of the proposed ContextLoRA against four baselines. (d), (e), and (f) show the robustness of our proposed ContextLoRA with different poisoned samples.}
\label{fig_performance}
\vspace{-4mm}
\end{figure*}

\begin{figure*}[!t]
\centering
\subfloat[]{\includegraphics[width=2.2in]{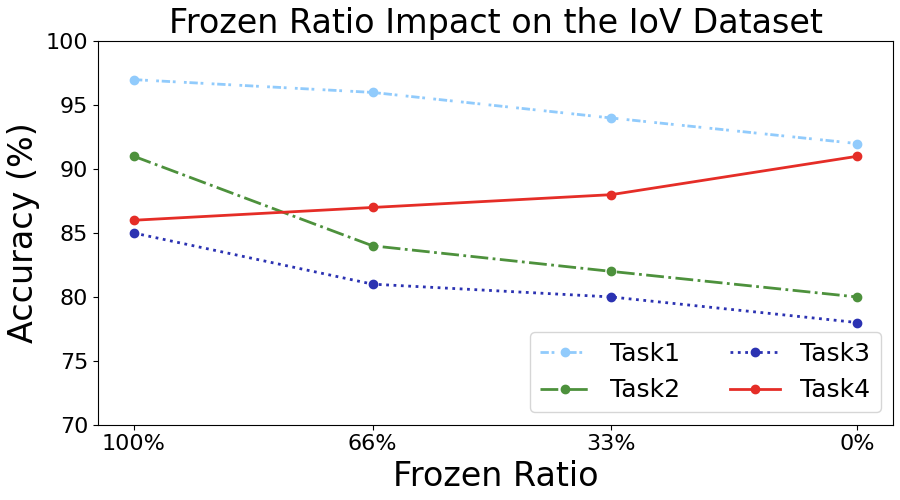}
\label{fig:8_first_case}}
\hfil
\subfloat[]{\includegraphics[width=2.2in]{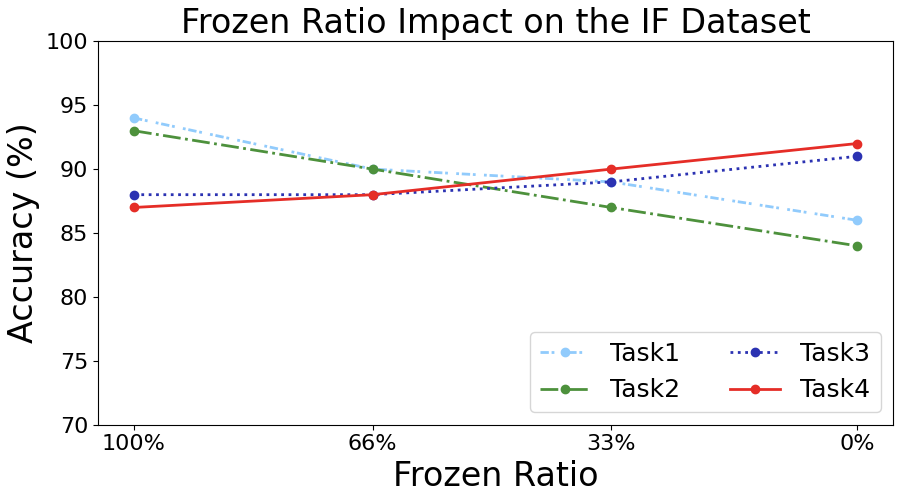}
\label{fig:8_second_case}}
\hfil
\subfloat[]{\includegraphics[width=2.2in]{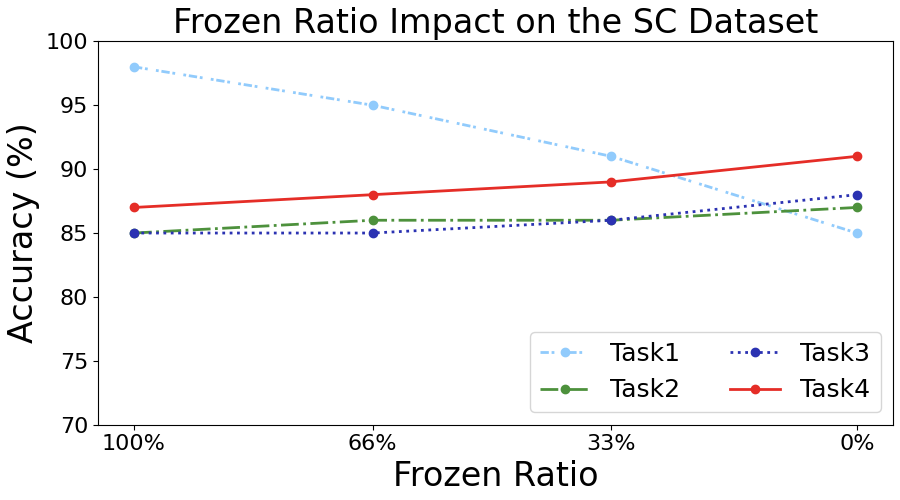}
\label{fig:8_third_case}}
\hfil
\subfloat[]{\includegraphics[width=2.2in]{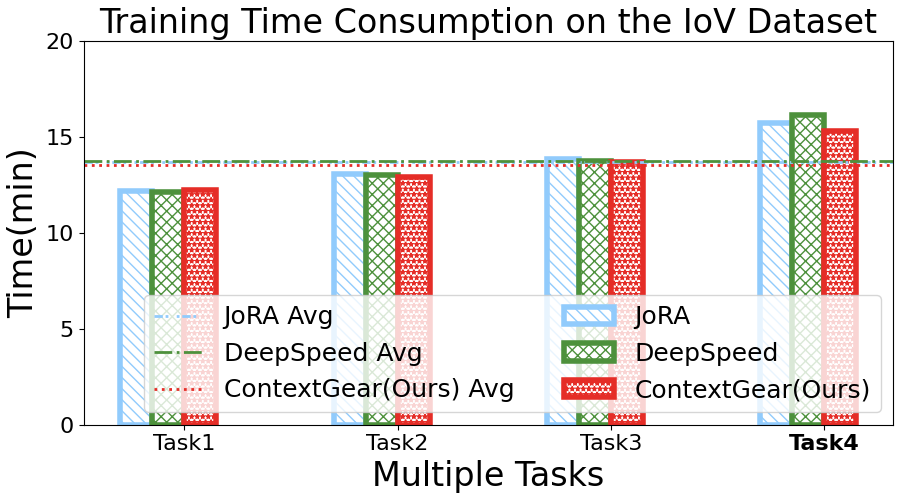}
\label{fig:8_fourth_case}}
\hfil
\subfloat[]{\includegraphics[width=2.2in]{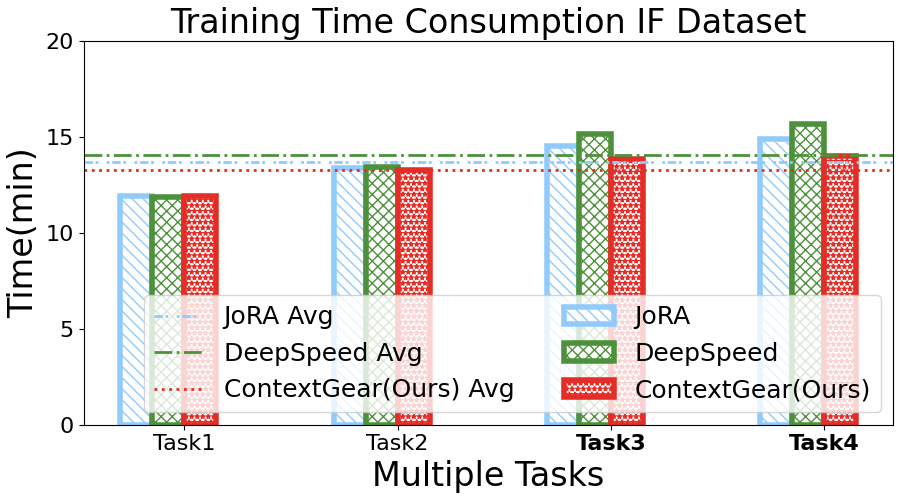}
\label{fig:8_fifth_case}}
\hfil
\subfloat[]{\includegraphics[width=2.2in]{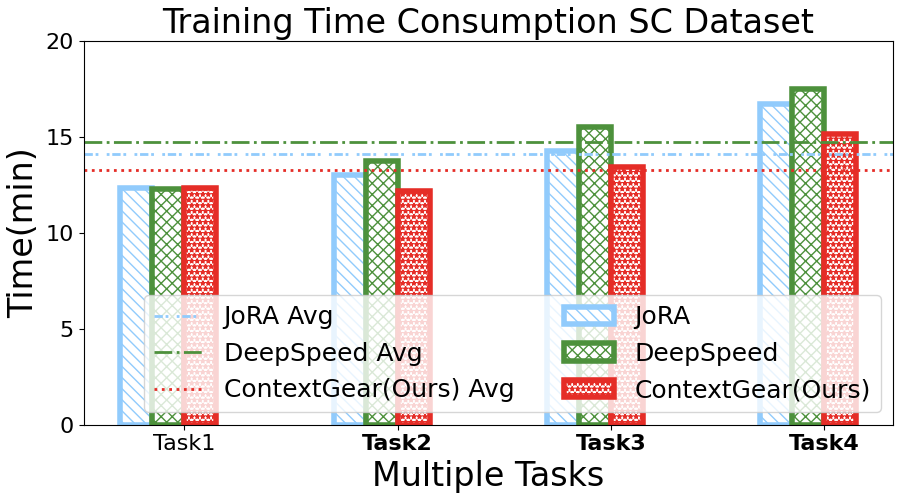}
\label{fig:8_sixth_case}}

\caption{Performance comparisons on three datasets. (a), (b), and (c) show the accuracy of multiple tasks under various frozen ratios. (d), (e), and (f) demonstrate the training time consumption of ContextLoRA after being accelerated by the proposed ContextGear.}
\label{fig_performance2}
\vspace{-4mm}
\end{figure*}

This section presents extensive experiments to evaluate the effectiveness of the proposed ContextLoRA and ContextGear methods, along with key observations and conclusions. 
Figure \ref{fig_performance} compares accuracy and robustness across three benchmarks, showing that ContextLoRA outperforms established baselines such as LoRA, HydraLoRA, and MoLE. Figures \ref{fig_performance2}\subref{fig:8_first_case}\subref{fig:8_second_case}\subref{fig:8_third_case} examine the impact of the frozen ratio on ContextLoRA’s performance, providing insights into balancing task priorities during training. Table \ref{tab:table2} summarizes ContextLoRA’s accuracy under simulated and real-world scenarios, confirming its robustness.
For ContextGear, Figures \ref{fig_performance2}\subref{fig:8_fourth_case}\subref{fig:8_fifth_case}\subref{fig:8_sixth_case} compare its training time efficiency with baselines such as JoRA and DeepSpeed, demonstrating significant improvements. Table \ref{tab:table3} further demonstrates how ContextGear accelerates fine-tuning under both simulated and real-world conditions.

\textbf{1) ContextLoRA}

\noindent
\textbf{Accuracy comparisons of the proposed ContextLoRA }: 
We evaluate our approach on three distinct datasets to assess its performance within a multi-task framework. The results demonstrate that our method achieves accuracy on the root task comparable to the multi-task LoRA baselines, while significantly surpassing the original LoRA approach. This suggests that training on subsequent tasks does not compromise the performance of the root task. Furthermore, for the subsequent tasks, our method significantly outperforms all baseline models. For example, as shown in Figure \ref{fig_performance}\subref{fig:7_first_case}, our approach achieves a notable high score of 93 on task 4, outperforming the baseline scores of 81 and 74. In the comparison of the IoV dataset, Task 1, Task 2, and Task 3 are independent tasks, so ContextLoRA maintains comparable performance over the baselines. In addition, for independent tasks within the task topology, our method often performs slightly lower than GPT-4o. However, for some subsequent complex tasks, such as Task 4 in Figures \ref{fig_performance}\subref{fig:7_first_case}\subref{fig:7_second_case}, our ContextLoRA, after being trained on LLaVA 7B, can even outperform GPT-4o. These experimental results highlight the ability of our approach to effectively capture complex, recursive relationships between tasks, thereby enhancing the model’s capacity for multi-task inference without sacrificing performance on individual tasks.

\begin{figure*}[!t]
\centering
\subfloat[]{\includegraphics[width=2.2in]{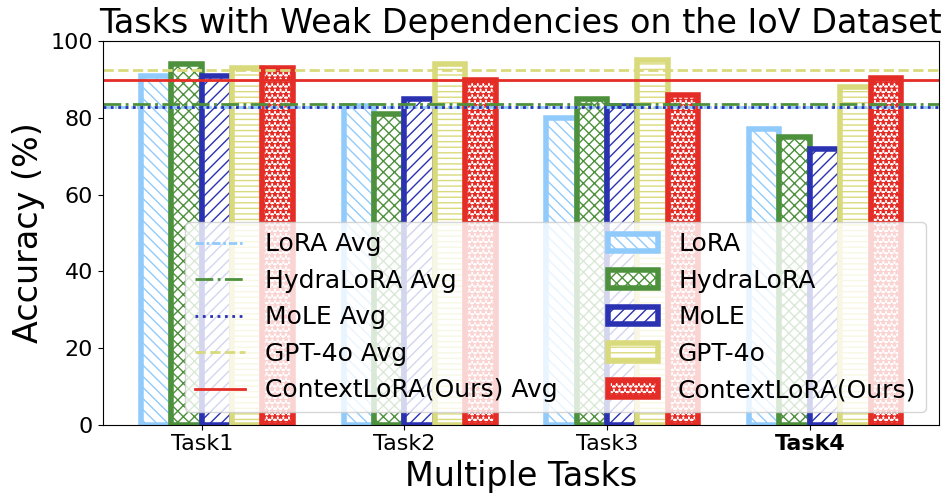}
\label{fig:9_first_case}}
\hfil
\subfloat[]{\includegraphics[width=2.2in]{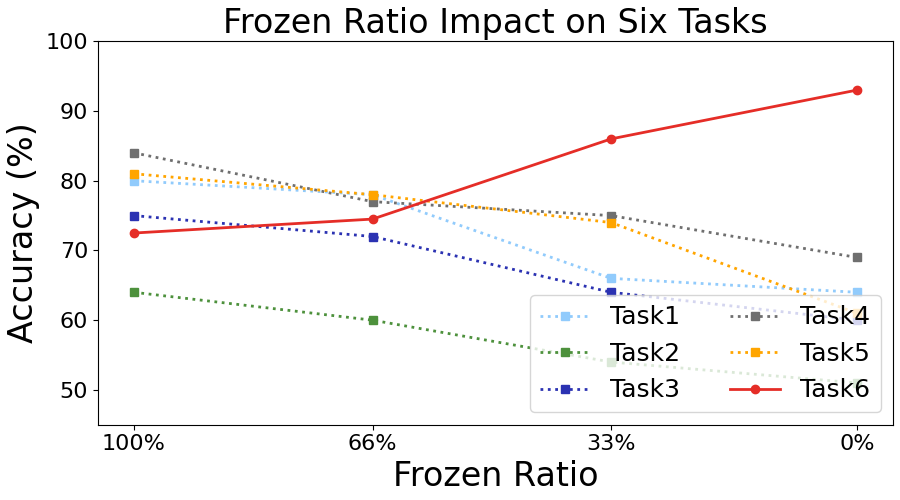}
\label{fig:9_second_case}}
\hfil
\subfloat[]{\includegraphics[width=2.2in]{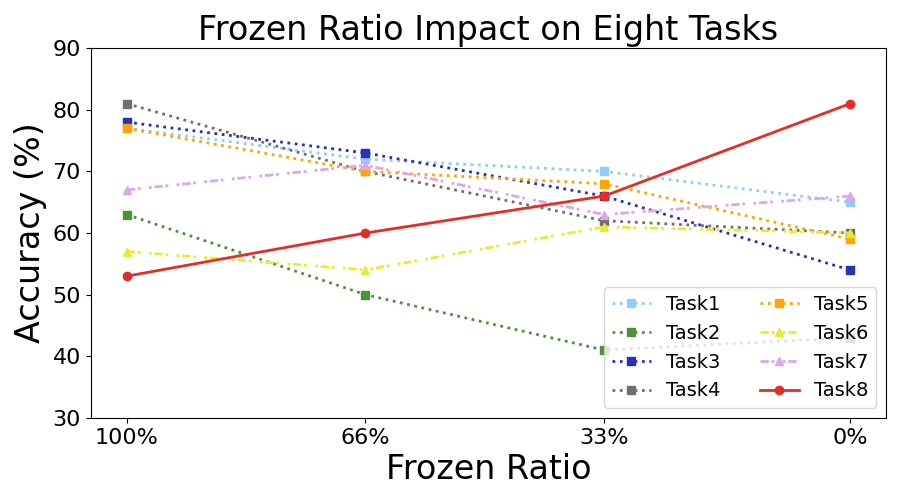}
\label{fig:9_third_case}}
\caption{Performance comparisons on the IoV datasets. (a) compare the accuracy of our proposed ContextLoRA against four baselines with weak task dependencies on the IoV dataset. (b) and (c) show the accuracy of our proposed ContextLoRA as the frozen ratio changes, with the number of tasks on the IoV dataset increased to six and eight.}
\label{fig9_performance}
\vspace{-4mm}
\end{figure*}

To explore the impact of task dependencies on the performance of ContextLoRA, we replace Tasks 1 and 2 in the IoV dataset with tasks that have weak dependencies on Task 4, in order to observe the changes in Task 4’s performance. As shown in Figure \ref{fig9_performance}\subref{fig:9_first_case}, the performance of Task 4 decreases slightly from 93\% to 90.5\%, but it still outperforms all baselines, with the highest baseline at 77\%.

\noindent
\textbf{Robustness of the proposed ContextLoRA}:
We evaluate the robustness of our method on Task 4 across three distinct datasets, each subjected to pollution levels of 0\%, 5\%, 10\%, and 20\%. We introduce controlled label noise by randomly shuffling QA pairs. We first sampled a fixed proportion of QA pairs from the training set, then independently reordered the questions and answers within this subset before recombining them at random to generate polluted examples. This perturbation was applied exclusively to basic, fixed-response questions, while open-ended queries remained intact. Notably, as illustrated in Figures~\ref{fig_performance}\subref{fig:7_fourth_case}\subref{fig:7_fifth_case}\subref{fig:7_sixth_case}, even at a 20\% pollution rate, our accuracy remains above 80\%, showing only a slight decrease compared to the unpolluted condition. In comparison to other baseline methods, our approach consistently outperforms them by at least 5\% across all pollution levels, with performance improvements exceeding 15\% in some cases. Furthermore, our method exhibits a significantly lower rate of performance degradation compared to these baselines. These results indicate that our method offers superior robustness under various levels of data pollution relative to other baseline approaches.

\noindent
\textbf{Our proposed ContextLoRA with different frozen ratio}: 
We conduct experiments on three distinct datasets, training the recursive module with varying freezing ratio $1-\delta$. As the freezing rate decreases, we observe a slight drop in accuracy for the parent node modules; for instance, as shown in Figure \ref{fig_performance2}\subref{fig:8_first_case}, Tasks 1, 2, and 3 all experience some degree of accuracy reduction with lower freezing rates. Conversely, accuracy for child nodes within the recursive module gradually improves with reduced freezing, as demonstrated by Tasks 3 and 4 in Figure \ref{fig_performance2}\subref{fig:8_second_case}, where performance increases as the freezing rate decreases. This trend is further corroborated by Figure \ref{fig_performance2}\subref{fig:8_third_case}, which highlights similar improvements in child-node performance. 

In addition, we also explore the impact of changes in the frozen ratio on the performance of our method as the number of tasks increases. As shown in Figures \ref{fig9_performance}\subref{fig:9_second_case}\subref{fig:9_third_case}, when the number of tasks expands to six and eight, we still observe the same phenomenon as in the four-task case. Specifically, for frozen tasks located earlier in the task topology, their performance declines as the freezing rate decreases. The frozen tasks near the end show certain fluctuations, while the performance of the training nodes at the end gradually increases as the frozen ratio decreases. 

These findings suggest that during recursive module training, adjusting the freezing rate can effectively balance task priorities, enabling the model to focus on specific tasks. This approach enhances the output capability of key progressive modules, while having a relatively small impact on the accuracy of other components.

\noindent
\textbf{Our proposed ContextLoRA in simulations and real-world tests}: 
We evaluate two versions of the LLaVA model, differing in parameter size, through simulation experiments, as well as the LLaVA 7B model deployed in a real-world system. As illustrated in Table~\ref{tab:table2}, when comparing models of similar structures, we observe that larger parameter sizes improve the model’s learning capacity, enabling it to fit more complex tasks, resulting in higher accuracy and enhanced learning performance. Additionally, we confirm that the model functions effectively in a real-world system, achieving results comparable to those observed in simulation experiments.

\begin{table}[!t]
\caption{Comparisons of Accuracy under Simulation and Real-world Test\label{tab:table2}}
\centering
\begin{tabular}{c|c c c c}
\hline
\ & \textbf{\begin{tabular}[c]{@{}c@{}}LoRA\end{tabular}}
& \textbf{\begin{tabular}[c]{@{}c@{}}HydraLoRA\end{tabular}} 
& \textbf{\begin{tabular}[c]{@{}c@{}}MoLE\end{tabular}} 
& \textbf{\begin{tabular}[c]{@{}c@{}}ContextLoRA(Ours)\end{tabular}} \\ 
\hline
{\begin{tabular}[c]{@{}c@{}}Fine-tuned \\ LLaVA 7B\end{tabular}} & 81\% & 74\% & 74\% & \textbf{86\%} \\
\hline
{\begin{tabular}[c]{@{}c@{}}Fine-tuned \\ LLaVA 13B\end{tabular}} & 83\% & 75\% & 77\% & \textbf{90\%}\\ 
\hline
{\begin{tabular}[c]{@{}c@{}}Real-world \\ tests\end{tabular}} & 79\% & 76\% & 74\% & \textbf{87\%}\\ 
\hline
\end{tabular}
\vspace{-3mm}
\end{table}

\textbf{2) ContextGear}

\noindent
\textbf{Time consumption of the proposed ContextGear compared with multiple baselines}:
ContextGear is specifically designed to optimize training within progressive modules, as illustrated in Figures \ref{fig_performance2}\subref{fig:8_fourth_case}\subref{fig:8_fifth_case}\subref{fig:8_sixth_case}. Across three different datasets, our approach demonstrates remarkable efficiency in training progressive modules, significantly outpacing the two baseline methods. Additionally, we maintain high efficiency in training the source node, with performance closely comparable to the baselines. The GPU memory consumption of the proposed ContextGear remains the same as that of the baseline LoRA methods. The substantial improvement in overall training speed can be attributed to the optimization introduced in our gradient backpropagation process.

\noindent
\textbf{Our proposed ContextGear under simulation and real-world tests}: 
We further demonstrate the advantages of ContextGear on two LLaVA models with different parameter sizes. As illustrated in Table~\ref{tab:table3}, on the lightweight LLaVA 7B model, our approach achieves significantly faster training speeds compared to the baseline model. As model size increases, the performance benefits of our optimizations become even more pronounced. Notably, for the LLaVA 13B model, when distributed across multiple nodes, ContextGear shows nearly a 20\% speed improvement over the best baseline. The superior performance of ContextGear in distributed systems is further highlighted in real-world system tests, where training speeds exceed one baseline by over 22\% and the other by more than 40\%. These results underscore the optimization capabilities of ContextGear, particularly in the contexts of multi-task learning and distributed training.

\begin{table}[!t]
\caption{Comparisons of Training Time Consumption under Simulation and Real-world\label{tab:table3}}
\centering
\begin{tabular}{c|c c c c}
\hline
\ & \textbf{\begin{tabular}[c]{@{}c@{}}JoRA\end{tabular}}
& \textbf{\begin{tabular}[c]{@{}c@{}}Deepspeed\end{tabular}} 
& \textbf{ContextGear(Ours)} \\ 
\hline
{\begin{tabular}[c]{@{}c@{}}Fine-tuned \\ LLaVA 7B\end{tabular}} & 15'45'' & 16'09'' & \textbf{15'19''} \\
\hline
{\begin{tabular}[c]{@{}c@{}}Fine-tuned \\ LLaVA 13B\end{tabular}} & 28'05'' & 29'35 & \textbf{24'34''}\\ 
\hline
{\begin{tabular}[c]{@{}c@{}}Real-world\\ tests\end{tabular}} & 39'42'' & 45'32'' & \textbf{32'27''}\\ 
\hline
\end{tabular}
\vspace{-3mm}
\end{table}

\begin{figure*}[!t]
\centering
\vspace{-4mm}
\includegraphics[width=1\textwidth,trim=20 100 30 155,clip]{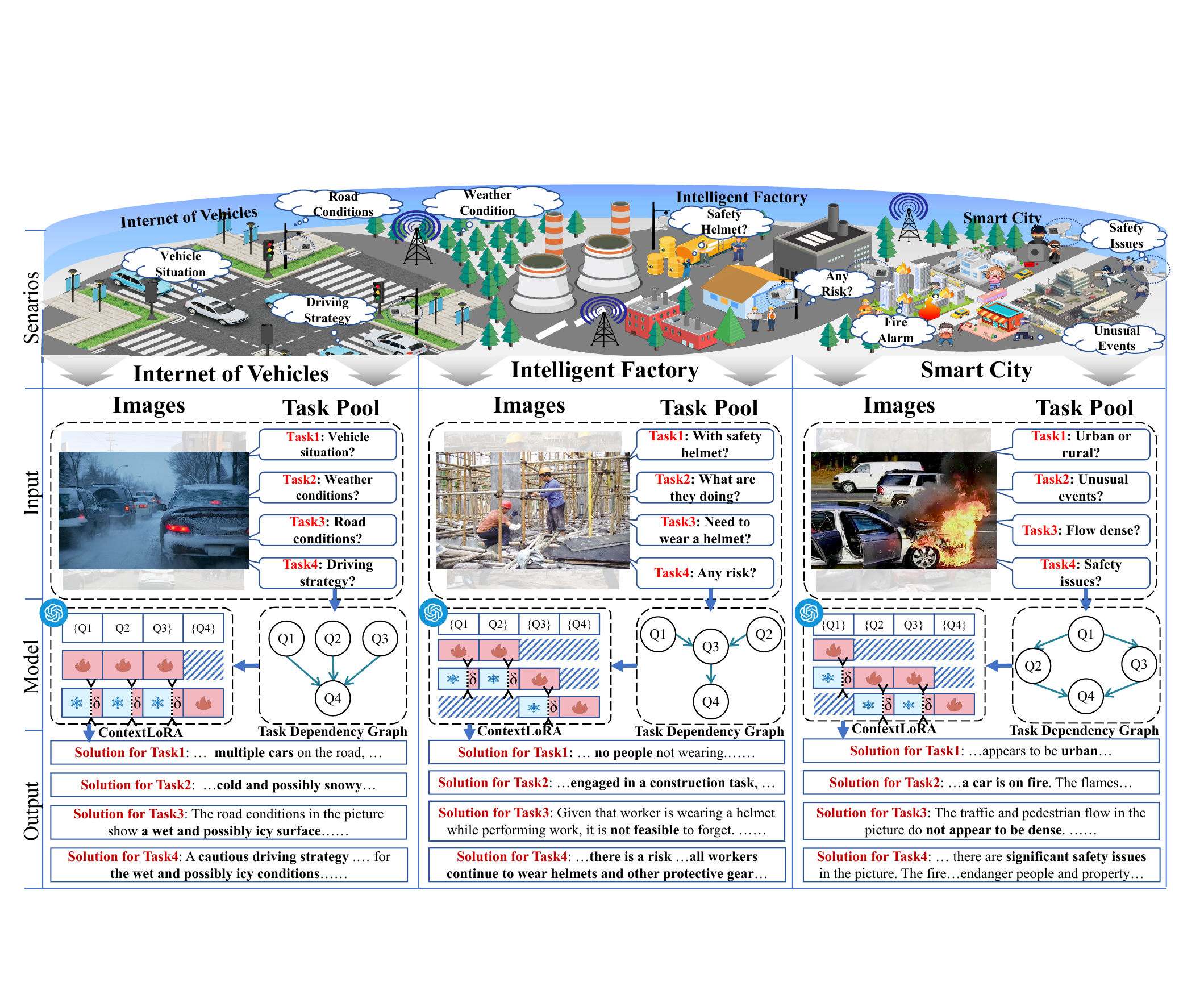}
\caption{Case study of the proposed ContextLoRA. At the top of this figure, three communication scenarios with multiple tasks are described. The workflow indicates that the vision inputs and multiple tasks are transmitted to multimodal LLMs equipped with our ContextLoRA. The output of LLMs is placed at the bottom, generating solutions for diverse computer vision applications in communication scenarios.}
\label{fig_case_study}
\vspace{-5mm}
\end{figure*}

\subsection{Case Study}

As illustrated in Figure \ref{fig_case_study}, we visualize three case studies to demonstrate the effectiveness of our proposed ContextLoRA algorithm on multiple tasks under the Internet of Vehicles, intelligent factories, and smart city scenarios. The system takes multimodal inputs in interactive scenarios along with multiple related tasks. These tasks are then transformed into a task dependency graph. The fine-tuned ContextLoRA model is used to generate the solutions for the respective tasks.

\subsubsection{\textbf{Internet of Vehicles}}Under the Internet of Vehicles scenario, we have formulated four interactive tasks. The first task to the third task aims to identify vehicle situation, weather conditions, and road conditions in the collected image, respectively. The fourth task caring about the driving strategy of a current environment, which is closely related, builds upon the first three tasks progressively. Taking this scenario as an example, we illustrate the overall process. Images and interactive tasks are inputs into the fine-tuned ContextLoRA multimodal LLM, which generates the final solution. For this specific image, the output provides a detailed assessment: several vehicles are parked along the roadside, the weather appears cold with possible snowfall, and the road surface is wet and icy. Consequently, the recommendation aligns with the image input, advising reduced speed and cautious driving.

\subsubsection{\textbf{Intelligent Factory}}Under the intelligent factory scenario, we have formulated four tasks, including helmet usage, worker activities, the necessity of helmets, and risk analysis. As shown in the task dependency graph in Figure \ref{fig_case_study}, the third task builds upon the first two tasks, and the ultimate fourth task is closely related to the third task. For the specific image, the output provides a detailed assessment: all workers are observed to be wearing safety helmets while performing construction tasks. Consequently, the model recommends all workers should continue using helmets to ensure safety in the construction environment.

\subsubsection{\textbf{Smart City}}Under the smart city security scenario, we have formulated four tasks, including identifying the environment, unusual events, the density of traffic flow, and safety issues. The fourth task builds upon the second and third tasks in a progressive manner. For this specific image, the output provides a detailed assessment: an unusual event is observed in an urban area with a car on fire, and the flow of traffic and pedestrians is not dense. Consequently, the output suggests that the fire will endanger people and property, and should be addressed immediately.

\section{Discussion}
\label{7Dis}
So far, we have shown the superiority of our proposed ContextLoRA on three benchmarks. In this section, we take a further step to highlight some interesting observations and future works, including the efficiency, robustness, and privacy of the proposed ContextLoRA.

\subsection{Our Proposed ContextLoRA for Real-time Applications}

We investigate the latency requirements of IMAs under several communication scenarios. For IoVs, traffic signal controlling task requires latencies on the level of seconds~\cite{tubaishat2008wireless, 8809662}. For industrial safety, remote monitoring demands latency between 100ms and 1s~\cite{zawish2022energy, 10412143}. Additionally, in urban security~\cite{liu2019edge, wang2017elastic,9139976}, second-level latency is required for face and vehicle recognition. In the simulation environment, the inference time of the proposed ContextLoRA is 0.487s/it for LLaVA 7B, while in a real-world testbed, it reaches 0.93s/it. Here, /it denotes a single-cycle iteration of processing a batch of data. These results confirm that our proposed ContextLoRA efficiently performs inference with limited computational resources and supports various real-time IMAs. Currently, achieving millisecond-level latency with LLMs remains a challenge. In the future, we plan to explore lightweight techniques to enable broader real-time applications for LLMs.

\subsection{Robustness of Our Proposed ContextLoRA}
Numerical results demonstrate that our proposed ContextLoRA significantly strengthens the robustness of the model by segmenting the LoRA parameter matrix. We confine potential issues to specific matrix segments, reducing the risk of widespread performance degradation. This architecture is particularly effective in scenarios involving data contamination or adversarial attacks, by localizing disruptions and preventing systemic failures. As illustrated in Figures~\ref{fig_performance}\subref{fig:7_fourth_case}\subref{fig:7_fifth_case}\subref{fig:7_sixth_case}, robustness experiments across three datasets with 20\% contamination show that our model maintains over 80\% accuracy, with minimal performance loss, at least 10\% higher than the baselines.

\subsection{Privacy of Our Proposed ContextLoRA}
To address privacy concerns raised by distributed training, recent studies~\cite{10666083, cho2023heterogeneous, Wu2024FedFMSL} have proposed combining federated learning with LoRA to enhance privacy protection. Our proposed ContextLoRA approach can similarly be integrated with federated learning, and we are looking forward to exploring it in future research.

\section{Conclusion}
\label{8Conclusion}
This paper studies compositional LLM reasoning with
structured task relations and the corresponding parallelism and optimization scheme. We present ContextLoRA and ContextGear to promote LLM applications for interactive multimodal communications. Specifically, the proposed ContextLoRA is a novel fine-tuning method aiming to efficiently learn multi-task dependency. Our proposed ContextGear is a parallelism and optimization algorithm accelerating ContextLoRA among edge devices. Towards the proposed ContextLoRA and ContextGear in practice, the design principles are dynamic, collaborative, interpretable, pluggable, robust, and efficient so that they can be applied to real-world scenarios. We conduct extensive experiments on three benchmarks with 12 tasks to show the effectiveness of the proposed ContextLoRA and ContextGear. We verify our methods on a real-world testbed proving the ability of LLM collaborative fine-tuning on edge devices. Future work includes privacy preservation in collaborative ContextLoRA training.

\bibliographystyle{IEEEtran}
\bibliography{reference.bib}
 
\vfill

\end{document}